\documentclass{article} 
\usepackage{iclr2024_conference,times}

\usepackage{amsmath,amsfonts,bm}

\def\eqref#1{equation~\ref{#1}}

\def\1{\bm{1}}

\DeclareMathAlphabet{\mathsfit}{\encodingdefault}{\sfdefault}{m}{sl}
\SetMathAlphabet{\mathsfit}{bold}{\encodingdefault}{\sfdefault}{bx}{n}

\usepackage[pagebackref=true,breaklinks=true,colorlinks,bookmarks=false]{hyperref}
\usepackage{url}
\usepackage{enumitem}
\usepackage{algorithm}
\usepackage{algorithmic}
\usepackage{colortbl}

\usepackage{subcaption,booktabs}
\usepackage{times}  
\usepackage{helvet}  
\usepackage{courier}  
\usepackage{graphicx} 
\usepackage{booktabs}
\usepackage{multirow, makecell}

\usepackage[most,breakable]{tcolorbox}
\usepackage{xcolor}
\usepackage{colortbl}
\usepackage{listings}
\newtcolorbox{mybox}{
    colback=blue!5, 
    colframe=blue!5, 
    rounded corners,
    sharpish corners,
    boxrule=0pt, 
    left=6pt,
    right=6pt,
    top=6pt,
    bottom=6pt,
}
\definecolor{bluegray}{rgb}{0.3, 0.38, 0.47}
\definecolor{whitesmoke}{rgb}{0.96, 0.96, 0.96}
\definecolor{codegreen}{rgb}{0,0.6,0}
\definecolor{codegray}{rgb}{0.5,0.5,0.5}
\definecolor{codepurple}{rgb}{0.58,0,0.82}
\definecolor{backcolour}{rgb}{0.96,0.96,0.94}

\lstdefinestyle{mystyle}{
    basicstyle=\ttfamily \lst@ifdisplaystyle\tiny\fi,
    breakatwhitespace=false,         
    breaklines=true,                 
    captionpos=b,                    
    keepspaces=true,                 
    numbers=left,                    
    numbersep=5pt,                  
    xleftmargin=12pt,
    showspaces=false,                
    showstringspaces=false,
    showtabs=false,                  
    tabsize=2,
    moredelim=[is][\bfseries]{<highlight>}{</highlight>}, %
    postbreak=\raisebox{0ex}[0ex][0ex]{\ensuremath{\color{black}\lst@ifdisplaystyle\hookrightarrow\fi\space}} %
}
\title{S-Agents: Self-organizing Agents in Open-ended Environments}

\author{Jiaqi Chen\footnotemark[1]
\quad Yuxian Jiang\thanks{These authors contributed equally to this work.}
\quad
Jiachen Lu
\quad
Li Zhang\thanks{Li Zhang (lizhangfd@fudan.edu.cn) is the corresponding author with School of Data Science, Fudan University.
}
\vspace{.2em} \\
Fudan University  \vspace{.6em} \\
\url{https://github.com/fudan-zvg/S-Agents}
}

\newcommand{\model}{S-Agents}

\iclrfinalcopy 
\begin{document}

\maketitle
\begin{figure*}[!h]
	\begin{center}
		\includegraphics[width=\linewidth]{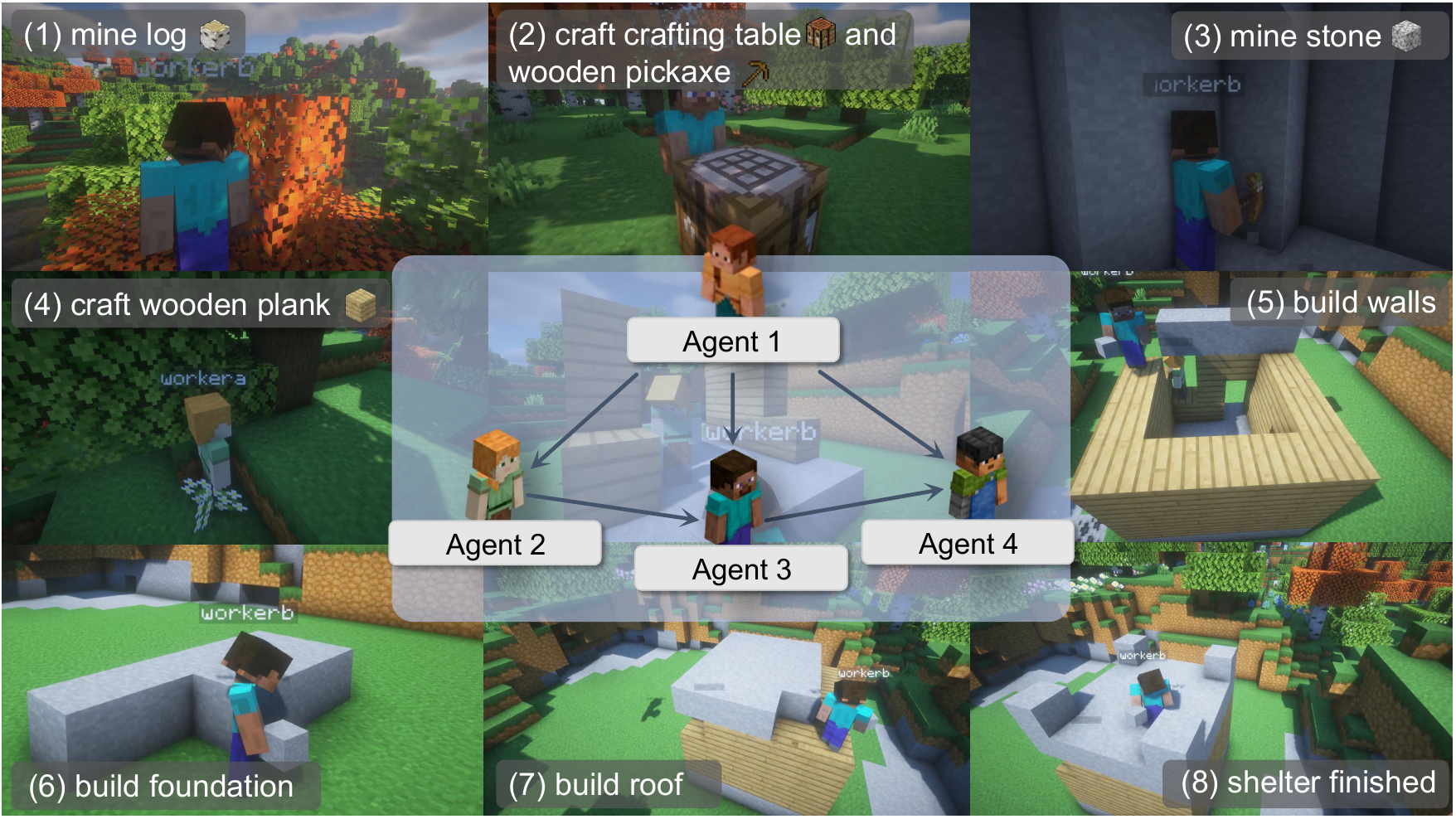}
	\end{center}
\vspace{-1em}
	\caption{
\textbf{Agent organization in open-ended environments.} Agent organization is a group of agents with a certain structure cooperating for shared goals. (1-3) depicts a group of agents collecting scattered rocks; (4-8) illustrates a group of agents building a shelter together. During their collaborative process, they autonomously orchestrated workflows without fixed steps by humans.}
 \vspace{-1em}
	\label{fig:cover}
\end{figure*}
\vspace{1em}

\begin{abstract}

Leveraging large language models (LLMs), autonomous agents have significantly improved, gaining the ability to handle a variety of tasks. In open-ended settings, optimizing collaboration for efficiency and effectiveness demands flexible adjustments. Despite this, current research mainly emphasizes fixed, task-oriented workflows and overlooks agent-centric organizational structures.
Drawing inspiration from human organizational behavior, we introduce a self-organizing agent system (\model{}) with a ``tree of agents" structure for dynamic workflow, an ``hourglass agent architecture" for balancing information priorities, and a ``non-obstructive collaboration" method to allow asynchronous task execution among agents. 
This structure can autonomously coordinate a group of agents, efficiently addressing the challenges of open and dynamic environments without human intervention.
Our experiments demonstrate that \model{} proficiently execute collaborative building tasks and resource collection in the Minecraft environment, validating their effectiveness. 

\end{abstract}  

\section{Introduction}
The fundamental objective of artificial intelligence has long been the development of intelligent autonomous agents with the capacity to operate proficiently in open-ended environments~\citep{weinbaum2017open,fujita2009intelligence}.
Autonomous agents powered by Large Language Models (LLMs)~\citep{kasneci2023chatgpt,touvron2023llama,ji2023masked}, especially GPT-4~\citep{openai2023gpt4} have paved the way for innovative developments in this domain. 
These models showcase remarkable competencies in diverse domains, including instruction comprehension~\citep{ouyang2022training,chung2022scaling}, decision-making~\citep{shinn2023reflexion,mandi2023roco,zhang2023building,yao2023tree}, tool usage~\cite{cai2023large}, code generation~\citep{hong2023metagpt,chen2023teaching}, and diverse other domains~\citep{tang2023medagents}. 
Moreover, LLM-powered agents  have been employed in diverse tasks in open-ended environments, spanning from sandbox games, simulated environments, and robotics~\citep{fan2022minedojo,zeng2022socratic,brohan2023rt,lu2023thinkbot,padmakumar2023multimodal}.
However, complex embodied tasks in open environments often necessitate collaboration among individual agents to achieve optimal outcomes~\citep{woolley2010evidence,fehr2000cooperation,zhuge2023mindstorms}. 

Unfortunately, the transition from single to multiple agents in an open-ended world introduces significant challenges in organizing a scalable group efficiently to tackle diverse tasks.
The principal challenge in multi-agent organizations resides in structuring a multitude of agents. 
Previous research has predominantly concentrated on the creation of \textit{\textbf{task-oriented, fixed workflows}}~\citep{hong2023metagpt, qian2023communicative,wu2023autogen}, neglecting to delve into the exploration of an organizational framework characterized by its \textit{\textbf{agent-centric approach and flexibility in workflow}}.
Identifying optimal connections among individuals and empowering agents to autonomously define a collective workflow present a novel challenge~\citep{grossi2006structural,jensen2017framework,chen2023scalable}.
Furthermore, agents in an organization must concurrently manage communication from both the surrounding environment and the organizational context.

\begin{figure}[t]
	\begin{center}
		\includegraphics[width=1.0\linewidth]{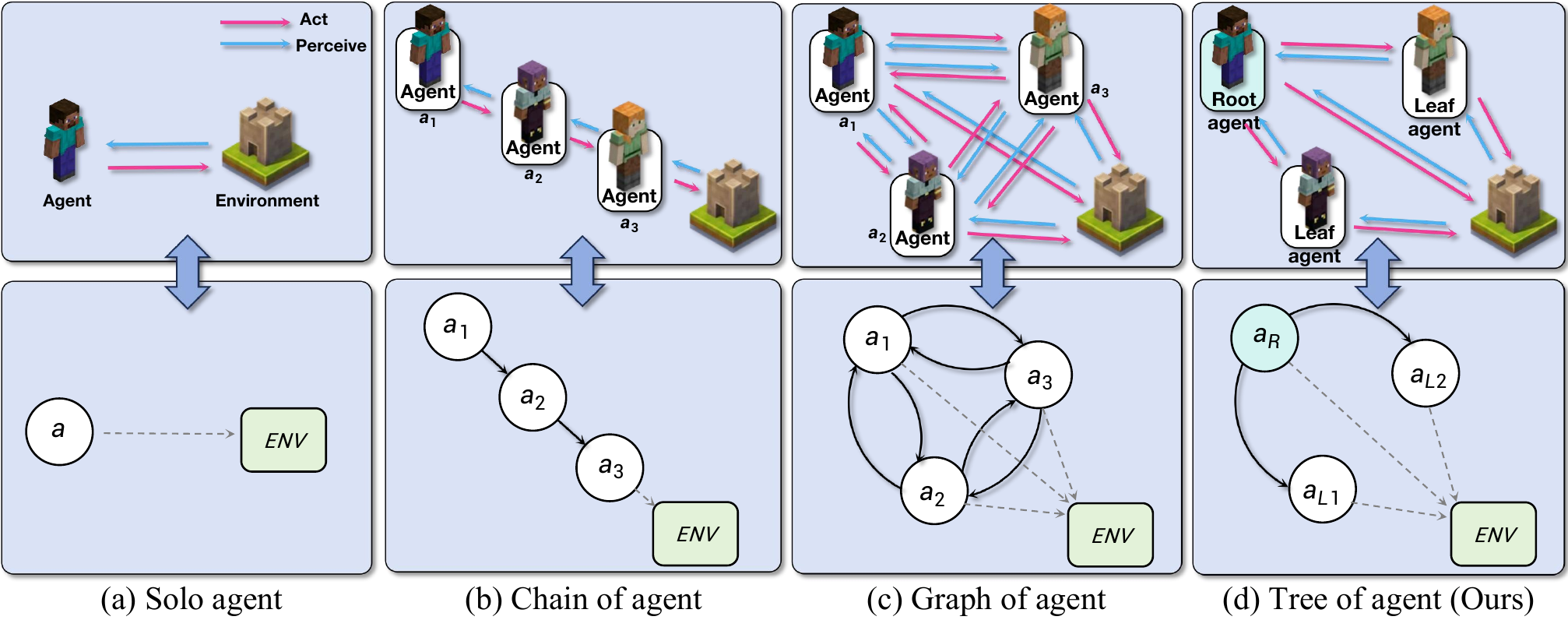}
	\end{center}
\vspace{-1em}
	\caption{\textbf{Schematic organizational structure comparison.} \textbf{(a)} Solo agent~\cite{wang2023voyager}: Direct interaction with the physical environment; \textbf{(b)} Chain of agents~\cite{qian2023communicative,hong2023metagpt}: Specialized agents sequentially perform their designated tasks and command the actions of the next agent; \textbf{(c)} Graph of agents~\cite{park2023generative}: Decentralized structure allowing all agents to command each other; \textbf{(d)} Tree of agents: Centralized structure retaining one agent as a leadership agent (root agent $a_{r}$), with other executor agents (leaf agent $a_{l1}, a_{l2}$) executing commands.}
	\label{fig:intra-org}
 \vspace{-1.5em}
\end{figure}

In this study, we introduce \model{}, a novel \textbf{\textit{self-organizing}} multi-agent system that allows agents to flexibly arrange workflow autonomously, without the need for predefined human instructions, in open-ended environments. The system is specifically designed to operate within the open-world game Minecraft as its environment.
It features:
1) A \textbf{\textit{``tree of agents"}} organizational structure, comprising a root node (leadership agent) and multiple leaf nodes (executor agents), illustrated in Figure~\ref{fig:intra-org}(d). The leadership agent autonomously arranges a flexible workflow without the need for human intervention.
2) An \textbf{\textit{hourglass agent architecture}} that strives to balance priorities between the agent community and the physical environment, promoting coordinated actions.
3) A \textbf{\textit{non-obstructive collaboration}} approach breaks away from the constraint of multiple intelligent agents sharing a fixed convergence beat, allowing agents to asynchronously execute collaborative tasks. This method is designed to alleviate delays induced by the slowest agent in each round, thereby enhancing overall efficiency.
\section{Related work}
\paragraph{LLM powered multi-agent collaboration}
Large Language Models (LLMs) have demonstrated phenomenal capabilities in various domains.
~\citet{park2023generative,akata2023playing,xiang2023language,gong2023mindagent} drive multiple agents with LLM and simulate human-like conversations as well as some social behaviors;
Recent attempts~\citep{li2023camel,dong2023self,qian2023communicative} found that multi-agent collaboration could develop software following a fixed process, but could not coordinate autonomously;
Multiple robotic arms~\citep{mandi2023roco}, and multiple agents~\citep{zhang2023building} working in collaboration all bring higher efficiency, but cannot be scaled up;
While previous methodologies have demonstrated leading-edge performance on certain tasks, their collaboration mechanisms are often preordained and task-centric.
However, the autonomous collaborative behavior of multi-agent embodied organizations in the open world remains an unclear topic in the current research.
We aim to design an agent-centric organization that empowers agents to directly orchestrate workflows, inherently determining their collaboration framework. This approach should be versatile enough to tackle a wide range of embodied tasks.

\paragraph{Embodied agents in Minecraft}
Minecraft is an open-ended, three-dimensional world that is a free experimental environment for building numerous benchmarks and agent methods.
\citet{fan2022minedojo,johnson2016malmo,kanervisto2022minerl} are established benchmarks for evaluating single-agent algorithms. 
While \citet{mohanty2022collecting,kiseleva2022interactive,kiseleva2022iglu,mohanty2023transforming} focus on designing specific structures based on human instructions.
On the other hand, Malmo~\citep{perez2019multi} offers an artificially designed game environment for multi-agent cooperation but lacks the necessary openness and diversity in tasks and environments.
Building upon these benchmarks, several advanced works explore different approaches to realizing embodied agents.
Many prior works utilize reinforcement learning to learn human game behavior~\citep{kanitscheider2021multi,lin2021juewu,mao2022seihai}.
\citet{fan2022minedojo,baker2022video} perform large-scale pre-training on game-playing videos.
\citet{wang2023describe,zhu2023ghost} proposes a closed-loop feedback framework for a single agent, allowing the agent to achieve its goals vis interact with the environment.
\citet{wang2023voyager} uses LLMs to automatically generate the next task based on the environment, continuously enriching the skill library and exploring the world.
\citet{zhao2023see} introduced visual perception capabilities to the LLM agent in Minecraft.
However, there has been limited exploration into the design of organizational structures for multiple LLM agents and the architecture of embodied agents that can be organized.

\section{Methodology}
In this section, we introduce LLM-based self-organizing agents (\model{}), including (i) an efficient directed tree of agents as an organizational structure, (ii) an hourglass agent framework for unified goal management and dynamic planning, and (iii) a non-obstructive collaboration paradigm allowing non-blocking parallelization.
\subsection{Organizational structure of agents}
\label{sec:agent_org}
\subsubsection{Agents as a graph}
We design an agent-centric organization, specifically, we do not predefine the specific roles and functions of agents~\citep{hong2023metagpt,qian2023communicative,wu2023autogen}. Instead, we place them in relationships, allowing them to autonomously allocate tasks and coordinate workflows based on circumstances and needs. This is the essence of \textit{\textbf{self-organizing}}.
Self-organizing agents collaborate to coordinate multiple sub-tasks \textbf{\textit{without the need for human intervention}}, working towards a shared goal.
This requires agents to have a certain organizational structure that allows tasks to be effectively transferred among agents.
To model the agent organizational structure, we formulate a graphical representation known as the {\bf\em agent graph} $\mathcal{G}=(\mathcal{V}, \mathcal{E})$. This graph is constructed based on the sets of agents, denoted as $\mathcal{A}= \{a_1, ..., a_n\}$, and the environment, represented as $p$,
\begin{equation}
    V=\left \{a_1,...,a_n,p\right \}, n > 1.
\end{equation}
We subsequently define the edges defined as \( \mathcal{E} = \{(e_{ij})\} \), where directed edge \( e_{ij} = (v_i, v_j) \) exists for \( v_i, v_j \in \mathcal{V} \), $i \neq j$ if \( v_i \) {actively acts} (takes actions on the environment $p$ or issues commands to other agents $\mathcal{V} - \{v_i, p\}$ and passively perceives feedback from \( v_j \).
The organizational structure of agents involves the design of the edge set $\mathcal{E}$ among agents. Existing research has primarily categorized these structures into the following two types.

\paragraph{Graph of agents}
The (fully connected) graph of agents~\citep{park2023generative} is an organizational structure where all agents are interconnected, allowing mutual command and feedback (shown in Figure~\ref{fig:intra-org}(c)), represented as
\begin{equation}
\mathcal{E}_{GoA} = \{(v_i, v_j) \mid v_i, v_j \in \mathcal{A}, i \neq j\}.
\end{equation}
Theoretically, this structure promotes extensive communication and facilitates the flow of tasks within agent group $\mathcal{A}$, contributing to the emergence of self-organizing properties.
However, interconnected agents (Sec.~\ref{exp:behavious}) exhibit bidirectional connections among agent pairs, \textit{i.e.}, $(a_i, a_j), (a_j, a_i) \in \mathcal{E}$, forming command cycles that allow for mutual command issuance.
These cycles introduce the potential for chaos, characterized by unpredictable and conflicting behaviors.

\paragraph{Chain of agents}
To address the issue of command cycles and capitalize on the expertise of specialized agents, a straightforward approach~\citep{hong2023metagpt} is the Chain of agents (CoA) (as in Figure~\ref{fig:intra-org}(b)), defined as follows:
\begin{equation}
\mathcal{E}_{CoA} = \{ {(v_i, v_{i+1})} { \mid v_i \in \mathcal{A}} \}.
\end{equation}
In this fixed structure, instructional information unidirectionally flows from the first agent to the last, culminating in the completion of the final product. 
For instance, in a task requiring stone excavation, the $a_1$ undertakes lumbering, passing the baton to the $a_2$ for crafting a wooden pickaxe, subsequently transmitting the task to $a_3$ for stone excavation.
While such a structure successfully avoids command cycles, it cannot flexibly adjust workflows, resulting in the potential for self-organization being compromised.

\begin{figure}[t]
	\begin{center}
		\includegraphics[width=1.0\linewidth]{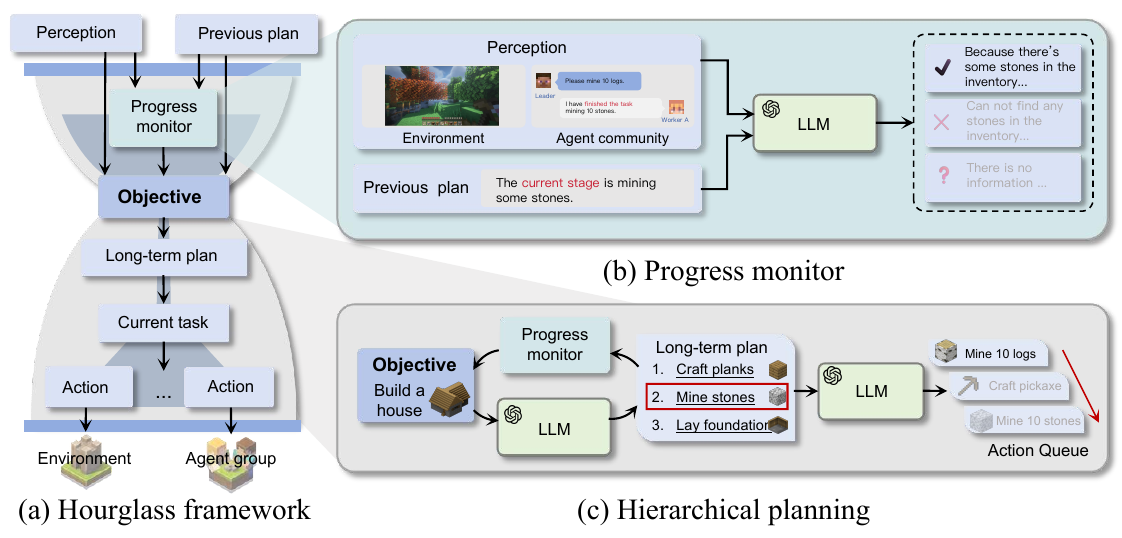}
	\end{center}
\vspace{-1em}
	\caption{
\textbf{An illustration of hourglass agent architecture.} \textbf{(a) \textit{Hourglass agent framework:}} 
\textit{The upper segment:} Processes inputs like perception and the previous plan. These inputs undergo a series of operations, converging towards a unified and consistent objective (the bottleneck of the hourglass). 
\textit{The lower segment:} Involves the decomposition of an objective through hierarchical planning. 
\textbf{(b) \textit{Progress monitor:}} Utilizes LLM to assess the current progress status of the ongoing task.
\textbf{(c) \textit{Hierarchical planning:}} Comprises two stages: Task planner and action planner. See Appendix~\ref{appendix:planning_process} and~\ref{sec:full_prompt_design} for example of planning and full prompt, respectively}
 \vspace{-1em}
	\label{fig:method}
\end{figure}

\subsubsection{Tree of agents}
To avoid the command cycles in GoA and maintain the potential for self-organization, we propose a directed {\bf\em tree of agents} (ToA), which introduces a leadership agent as the root node of the agent tree, with other agents serving as leaf nodes, as illustrated in Figure~\ref{fig:intra-org}(d).
Let $\mathcal{V} = \{a_r, a_{l1}, a_{l2}, ..., a_{ln}, p\}$, where $n > 2$, $a_r$ denotes root agent, and $a_{l}$ denotes $i$-th leaf agent. Define the edges $\mathcal{E}$ as:
\begin{equation}
   \mathcal{E}_{ToA} = \mathcal{E}_{\text{root}} \cup \bigcup_{i=1}^{n-1} \mathcal{E}_{li}.
\end{equation}
In this structure, the root agent $a_r$ serves as the central command authority, directing tasks to the leaf agents, $\mathcal{E}_{\text{root}} = \{(a_r, a_i) \mid a_i \in \{a_{l1}, a_{l2}, ..., a_{ln}\}\}$. Leaf agents $a_{l}$ interact with the environment to execute assigned tasks and do not actively command other agents. Therefore, $\mathcal{E}_{li}$ is an empty set.
Note that, leaf agents can communicate but not command each other.
This hierarchical approach ensures a clear flow of commands, avoiding the issues associated with command cycles and allowing for efficient task execution.
Furthermore, upon closer examination, we observed that agents experiencing simultaneous directives from multiple counterparts often result in conflicting and chaotic behaviors. 
Consequently, we mandate that the in-degree of each agent be restricted to less than \(1\), distinguishing it from the fully connected graph of the agent.

\subsection{Hourglass agent architecture}
\label{sec:hourglass_pivot}
In the organizational context, agents simultaneously perceive messages from the agent group $\mathcal{A}$ and information from the physical environment $p$. For instance, a leadership agent directs $a_l$ to mine iron, but $a_l$ is currently under zombie attack. 
The \textit{\textbf{duality of inputs}} poses a challenge for pure LLM decision-making, making it difficult to generate \textit{\textbf{consistent and reliable behavior}}.
To address this challenge, we propose the hourglass agent architecture (see Figure~\ref{fig:method}(a)). This framework filters the abundant information to distill a singular objective as a bottleneck. Subsequently, it decomposes this objective into a long-term plan and generates an executable actions queue as output.

\begin{figure*}[t]
	\begin{center}
		\includegraphics[width=1.0\linewidth]{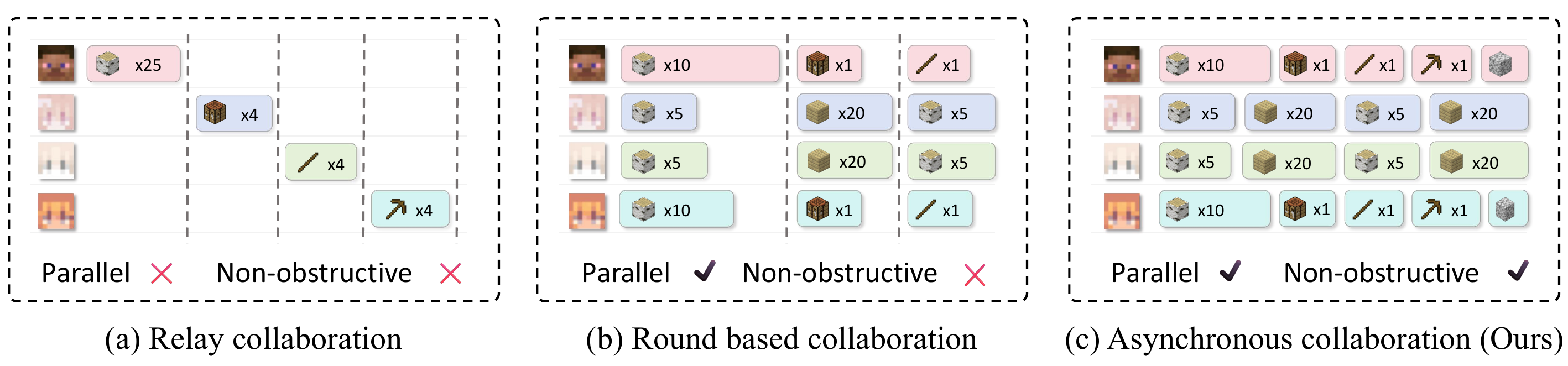}
	\end{center}
\vspace{-1em}
	\caption{
\textbf{Comparison of collaboration strategies.} (a) involves one agent sequentially executing tasks after another, with no parallelization; (b) is round-based, executing round by round, while (c) is asynchronous. \texttt{Colored regions} indicate tasks being performed, and \texttt{white regions} denote agent idleness.}
 \vspace{-1em}
	\label{fig:rollout}
\end{figure*}

\subsubsection{Perception: from agent group and physical environment}
\label{sec:Perception}
The perception module integrates feedback from the physical environment and dialogue transcripts from the agent group. 
\textbf{\textit{1) Physical environment:}}
The physical environment ${p}$ furnishes a diverse set of data, encompassing the \texttt{inventory}, \texttt{equipment}, and \texttt{nearby blocks}, \texttt{biome}, \texttt{time}, and \texttt{health and hunger bars}, and \texttt{3D coordinate} and more.
This data structure aligns with the one utilized in Voyager~\citep{wang2023voyager}.
\textbf{\textit{2) Agent group:}}
Utilizing language as an interface for communication within the agent group $\mathcal{A}$, we meticulously record the interactions initiated by the current agent. Each record includes  including \texttt{time}, \texttt{speaker}, \texttt{respondent}, and \texttt{message}. 

\subsubsection{Progress monitor}
\label{sec:progress_monitor}

The progress monitor, utilizing an LLM for evaluation, takes various perceptual information and the previous plan as input, generating the current task's completion status (``success", ``fail", or ``ongoing") along with its rationale. This evaluation occurs when there are no immediate pending actions.
As shown in Figure~\ref{fig:method}(b), the presence of stones in the inventory signals the completion of the mining task. The rationale for this determination is the sufficient quantity of stones already acquired.
For collaborative tasks, the assessment results should be based on the communication within the agent group.
\subsubsection{Hierarchical planning}
\label{sec:hierachical_planning}

As illustrated in Figure~\ref{fig:method}(c), a hierarchical planner involves the 
\textit{\textbf{two-step decomposition}} of a high-level objective, which can be broadly divided into two LLM-driven modules: task planner and action planner.
\textbf{\textit{1) Task planner}:}
As shown in Figure~\ref{fig:method}(c), the task planner, following the Chain-of-Thought (CoT) principles~\citep{wei2022chain}, employs a LLM for objective analysis and long-term planning.
The phase also includes choosing the immediate task for execution, marked as \texttt{current task} in the mint-colored module of Figure~\ref{fig:detail} in the Appendix.
\textbf{\textit{2) Action planner:}}
As depicted in Figure~\ref{fig:method}(c), the action planner accepts the current task as input and utilizes LLM to produce a sequence of executable actions, collectively known as an action queue (highlighted in green in Figure~\ref{fig:detail} in the Appendix). 
These actions are categorized into two types: \textit{\textbf{direct execution action}}, typically comprising an action, an object, and an optional location (e.g., ``Craft [quantity] [item] at [position]"), and \textit{\textbf{delegation action}}, where tasks are assigned to another agent (e.g., ``Instruct [player] to eliminate [quantity] [mob] at [position]").
Each action follows the approach of ~\citet{wang2023voyager}, referencing the most similar skill in the library, generating JavaScript code, and executing it in the Minecraft environment.
Each action in the queue is dequeued and executed sequentially until the queue is empty. Upon depletion of the queue, the progress monitor collects all perceptual information acquired during the execution of these actions to evaluate the task's completion status, as detailed in Sec.~\ref{sec:progress_monitor}.
\begin{figure*}[t]
	\begin{center}
		\includegraphics[width=1.0\linewidth]{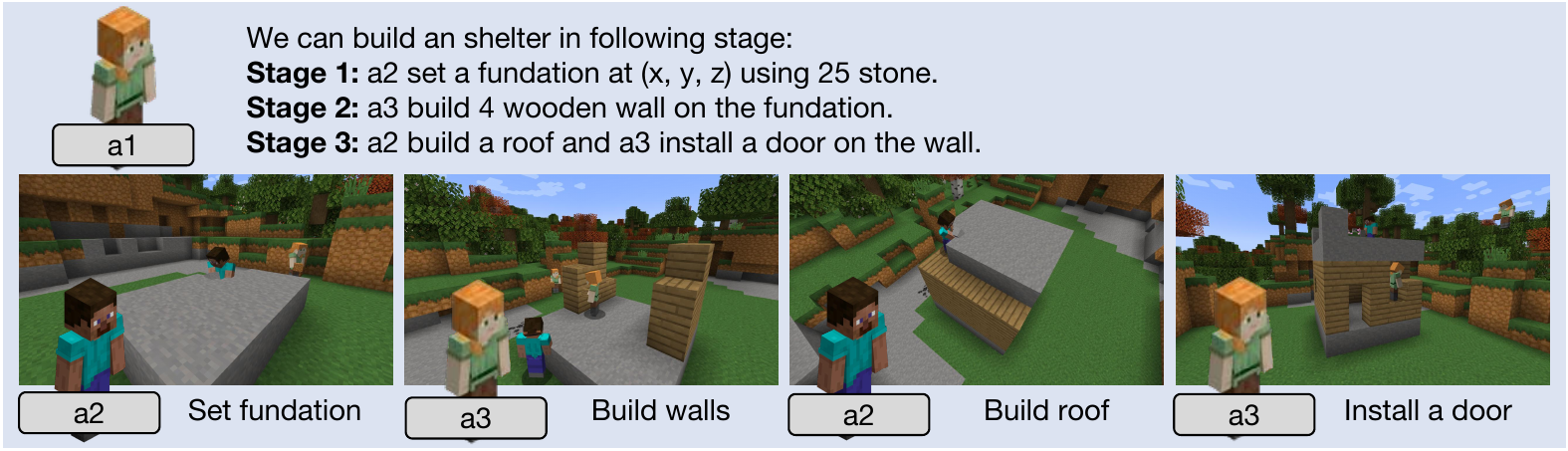}
	\end{center}
	\vspace{-0.5em}
	\caption{\textbf{Collective shelter construction.} The root agent (a1) systematically arranges the tasks and schedules the leaf agents (a2 \& a3) for phased execution. More screenshots can be referenced in Appendix~\ref{appendix:execution_process}.}
	\vspace{-1em}
	\label{fig:shelter}
\end{figure*}
\begin{figure*}[t]
	\begin{center}
		\includegraphics[width=1.0\linewidth]{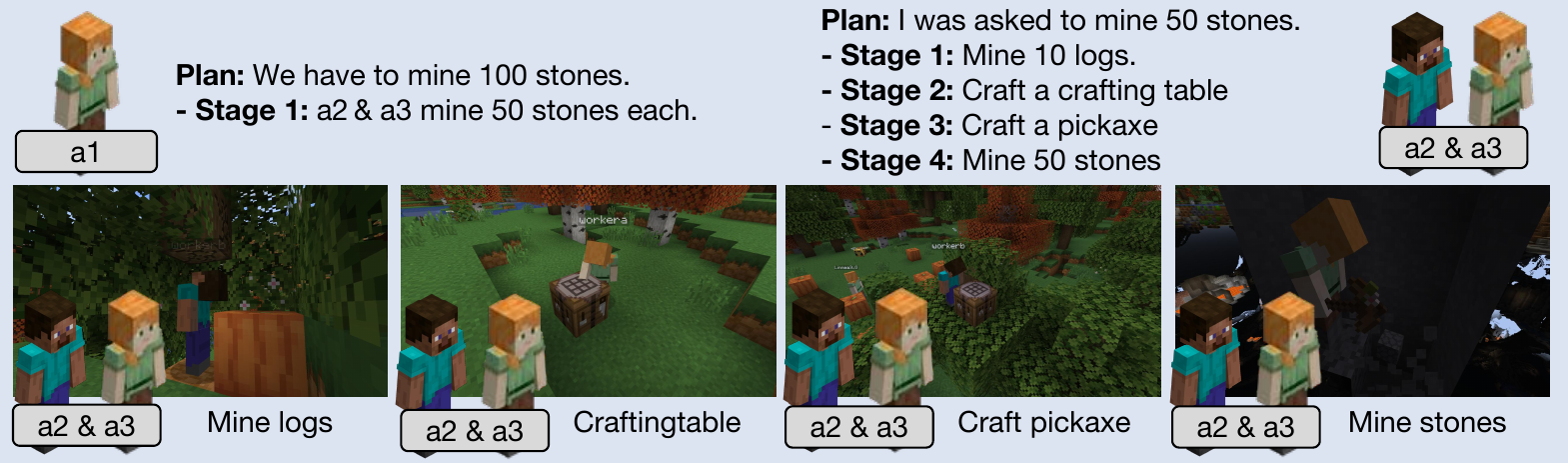}
	\end{center}
	\vspace{-0.5em}
	\caption{\textbf{Collective collection (mine 100 stones).} The root agent (a1) allocates tasks to facilitate parallel processing, while the leaf agent (a2 \& a3) autonomously refines these tasks into actionable steps. More screenshots can be referenced in Appendix~\ref{appendix:execution_process}}
	\vspace{-2em}
	\label{fig:mine_stones}
\end{figure*}
\subsection{Non-obstructive collaboration}
\label{sec:non_obstructive_decision_making}
The previous approaches can be categorized into two distinct types. 
In relay collaboration, exemplified by Chain of agents~\citep{hong2023metagpt} as illustrated in Figure~\ref{fig:rollout}(a), one agent initiates only after the completion of another, resulting in a sequential progression. This approach typically leads to a sequential and fully interdependent task execution among agents.
Relay collaboration~\citep{chen2023agentverse}, depicted in Figure~\ref{fig:rollout}(b), involves the simultaneous operation of all agents. Subsequently, it aggregates the outcomes of all agents after each round to inform task allocation for the subsequent round. Specifically, outside the individual agents, there is an outer \texttt{for-loop} that controls the round.
While this method introduces parallelism, it still introduces a bottleneck by impeding the slowest agent in each round, thereby constraining overall efficiency.
To address these limitations, we propose a non-obstructive asynchronous collaboration paradigm (illustrated in Figure~\ref{fig:rollout}(c)), where each agent operates independently. Once they complete their tasks, they directly report to the root agent to receive instructions for the next steps. Technically, we model each agent as an \textit{\textbf{independent asynchronous process}} that shares a message pool for communication. 
\section{Experiments}
\subsection{Experimental setup}
We evaluate our self-organizing agents on their collective collection performance (in Figure~\ref{fig:mine_stones}) and collective shelter construction capability (in Figure~\ref{fig:shelter}).

\textbf{\textit{Collective collection:}} 
Agents in the group must gather a specified amount of basic resources like wood, stone, and iron. 
They start without equipment and inherit the pre-trained skill library from Voyager~\citep{wang2023voyager}.
In contrast to completing the exploration of a single item~\citep{wang2023voyager,wang2023jarvis1,zhu2023ghost}, which entails gathering a diverse but limited quantity of items, this form of exploratory task can be autonomously executed by a single agent. The difficulty escalates when amassing a substantial quantity of resources, requiring agents to navigate numerous open-ended world events.
The considerable workload emphasizes the preference for a division of labor.

\textbf{\textit{Collective shelter construction:}}
For a basic shelter, agents need a stone foundation, wooden walls, and a stone ceiling. 
Two agents start with wooden planks, one with stone. This task assesses the leadership agent's \textit{\textbf{task assignment}} and challenges the consideration of \textit{\textbf{task dependencies}}. 
Effective progress management, an unexplored aspect in prior research, is crucial for successful, staged construction.

\paragraph{Metrics} We utilize the subsequent metrics:
\textbf{\textit{Time cost (TC):}} the time resources needed to execute a task. Reduced time expenditure suggests greater system efficiency. \texttt{NaN} denotes no progress for over 40 minutes when attempts exceed 5 without success.
\textbf{\textit{Mean prompt times (mPT):}} the average times of hierarchical planning iterations of each agent.

\paragraph{Implement details} 
We use Minecraft 1.19 with Fabric 0.14.21 as our testing environment. 
For large language models, we leverage OpenAI’s gpt-4~\citep{openai2023gpt4} for planning of root agent and task evaluation. 
Additionally, we utilize the gpt-3.5-turbo-16k APIs~\citep{introduce2022chatgpt} in all other settings. 
For text embedding, we leverage the capabilities of the text-embedding-ada-002 API. 
We configure all temperatures to 0, providing the best-fitted outcomes, and use temperature = 0.9 to encourage chatting among agents.
Following Voyager~\citep{wang2023voyager}, our simulation environment is constructed on the MineDojo framework~\citep{fan2022minedojo}, and we make use of the Mineflayer JavaScript APIs~\citep{PrismarineJS} for motor controls. We initialize all agents using the skill library pre-trained by Voyager~\citep{wang2023voyager}. 

\subsection{Evaluation Results}
\subsubsection{Main results}

\textit{\textbf{Impact of organizational structure:}}
In Table~\ref{tab:performance}(a), chain of agents (CoA) involves one agent sequentially executing $1/3$ collection tasks after another, with no parallelization. Graph of agents (GoA) avoids command cycle issues due to the upgrade to the more powerful gpt-4. Tree of agents (ToA) achieves the shortest completion time for identical tasks, requiring only 7.5 minutes and 3.8 mPT.

\begin{table*}[t]
\vspace{-0.2cm}
\renewcommand\tabcolsep{6pt}
\renewcommand\arraystretch{1.2}
\small
\begin{subtable}[t]{0.32\linewidth}
\centering
\begin{tabular}{l|cc>{\columncolor[gray]{.9}}c}
\hline

\hline

\hline

\hline
Structure& CoA & GoA & ToA \\
\hline

\hline
\hline
 TC & 29.0 & 9.3 & \textbf{7.5} \\
 mPT & 4.0 & \textbf{3.0} & 3.8 \\
\hline

\hline

\hline

\hline
\end{tabular}
\caption{
\textbf{Organizational structure.} Task: mining 50 stones.
CoA, GoA and ToA each consist of 3 worker agents.
%
}
\label{tab:misalignment}
\end{subtable}
\hspace{0.5cm}
\begin{subtable}[t]{0.65\linewidth}
\centering
\begin{tabular}{l|cccc}
\hline

\hline

\hline

\hline
Type & 50 logs & 100 logs & 50 irons & 100 irons \\
\hline

\hline
\hline
Solo agent & 10.2 & 18.3 & NaN & NaN \\
\rowcolor[gray]{.9} 
Agent organization & \textbf{5.1} & \textbf{11.1} & \textbf{15.3} & \textbf{24.3} \\
\hline

\hline

\hline

\hline
\end{tabular}
\caption{
\textbf{Agent organization \textit{vs.} Solo agent.}
Efficiency comparison (time cost/min) between agent organization and solo agent across different task difficulties. \texttt{NaN} denotes no progress for over 40 minutes. 
}
\label{tab:corruption}
\end{subtable}
\vspace{-0.3cm}
\caption{
\textbf{Performance of our agent organization on collective collection.}
}
\label{tab:performance}
\vspace{-0.5cm}
\end{table*}
\begin{table*}[ht]
\renewcommand\tabcolsep{7.5pt}
\renewcommand\arraystretch{1.2}
\small
\begin{subtable}[t]{0.4\linewidth}
\centering
\begin{tabular}{l|ccc}
\hline

\hline

\hline

\hline
N & 50 logs & 50 stones & 50 irons \\
\hline

\hline
\hline
1  & 10.2 & \textbf{6.5} & NaN\\
2  & 7.1 & 6.9 & NaN\\
3  & 8.5 & 8.6 & NaN\\
\rowcolor[gray]{.9} 
4  & \textbf{5.1} & 7.5 & \textbf{15.3}\\
\hline

\hline

\hline

\hline
\end{tabular}
\caption{
\textbf{Scale of agent organization.}
Metric: Time cost (/min).
``\texttt{N}" represents the number of agents.
In systems with more than one agent, we adopt a ToA group, which consists of 1 root agent and (N-1) leaf agents.
}
\label{tab:ablation}
\end{subtable}
\hspace{0.1cm}
\begin{subtable}[t]{0.57\linewidth}
\centering
\begin{tabular}{l|c>{\columncolor[gray]{.9}}c|c>{\columncolor[gray]{.9}}c}
\hline

\hline

\hline

\hline
Setting & \multicolumn{2}{c|}{\underline{Single agent}} & \multicolumn{2}{c}{\underline{3-agent organization}}  \\
Model & Voayer & Hourglass & Voayer& Hourglass \\
\hline

\hline
\hline
50 logs & 11.5 & \textbf{10.2} & - & \textbf{5.1}\\
50 stones & 6.9 & \textbf{6.5} & - & \textbf{7.5} \\
50 irons & NaN & NaN & - & \textbf{15.3}\\
\hline

\hline

\hline

\hline
\end{tabular}

\caption{
\textbf{Hourglass framework \textit{vs.} Voayer (baseline).}
The time cost (/min) of different architectures during the execution of collection tasks under a single agent and a 3-agent organization setting.
``-" indicates that Voyager does not support multi-agent communication and cooperation.\\
}
\label{tab:corruption}
\end{subtable}

\begin{subtable}[t]{0.2\linewidth}
\centering
\begin{tabular}{l|cc}
\end{tabular}

\end{subtable}
\begin{subtable}[t]{0.6\linewidth}
\centering
\begin{tabular}{l|cc}
\hline

\hline

\hline

\hline
Collaboration strategy & Time cost & Mean prompt time\\
\hline

\hline
\hline
Obstructive & 29.0 & 7.5 \\
\rowcolor[gray]{.9} 
Non-obstructive  & \textbf{4.0} & \textbf{3.8} \\
\hline

\hline

\hline

\hline
\end{tabular}

\caption{
\textbf{Comparison of collaboration strategies.} 
Tested a 3-agent system with different collaboration modes on the ``50 stone" task.
}
\label{tab:corruption}
\end{subtable}
\vspace{-0.2cm}
\caption{
\textbf{Ablation Study.}
}
\vspace{-0.8cm}
\label{tab:ablation}
\end{table*}

\textit{\textbf{Multi-agent organization \textit{vs.} Solo agent:}}
As shown in Table~\ref{tab:performance}(b), for simpler tasks like mining 50/100 logs, employing multiple agents significantly reduces the time by 5.1 and 7.2 minutes, respectively. 
For more challenging tasks (50/100 irons), solo agent attempts were largely ineffective (indicated as \texttt{NaN} in Table~\ref{tab:performance}(b)). 
A solo agent faces heightened probabilities of encountering various unforeseen challenges in an open-world setting, such as getting lost, obstructions by surrounding blocks, prolonged execution times, game environment exits, or difficulty finding iron. 
In the ToA system, despite all leaf agents making concerted efforts, once a leaf agent successfully mines iron, the root agent efficiently delegates the entire task to that agent, persisting until success. 

\subsubsection{Ablation study}
\textit{\textbf{Ablation on organization scale:}}
As shown in Table~\ref{tab:ablation}(a), although collecting wood is simple and repetitive, it requires agents to search for trees in the vicinity. 
Multi-agent collaboration can reduce the number of searches each person makes, and improve team efficiency. 
In the search for hard-to-find iron ore, multi-agent operations have increased the probability of finding ore sites, and a successful discovery can meet the demand. 
Meanwhile, because the location of stone resources is widespread and only requires digging 2-3 layers, the efficiency of a single agent is already high; the overall time cost of multi-agent collaboration depends on the time taken by the slowest agent.

\textit{\textbf{Hourglass framework vs. Voayer:}}
Table~\ref{tab:ablation}(b) shows our comprehensive advantages in the collection task. Not only do we perform faster in solo agent collection tasks compared to our baseline Voyager~\citep{wang2023voyager}, but we also enhance efficiency through supporting team collaboration, which is crucial for the future of scaled intelligent agents.

\textit{\textbf{Ablation on non-obstructive collaboration:}}
In the experiments described in Table~\ref{tab:ablation}(c), we employed relay collaboration to demonstrate obstructive strategy, while the asynchronous paradigm was used to illustrate the effects of non-obstructive approaches (see Figure~\ref{fig:rollout}). The experimental results show that non-obstructive collaboration reduces the time cost to 6.25 times the obstructive strategy.

\begin{figure*}[t]
	\begin{center}
		\includegraphics[width=1.0\linewidth]{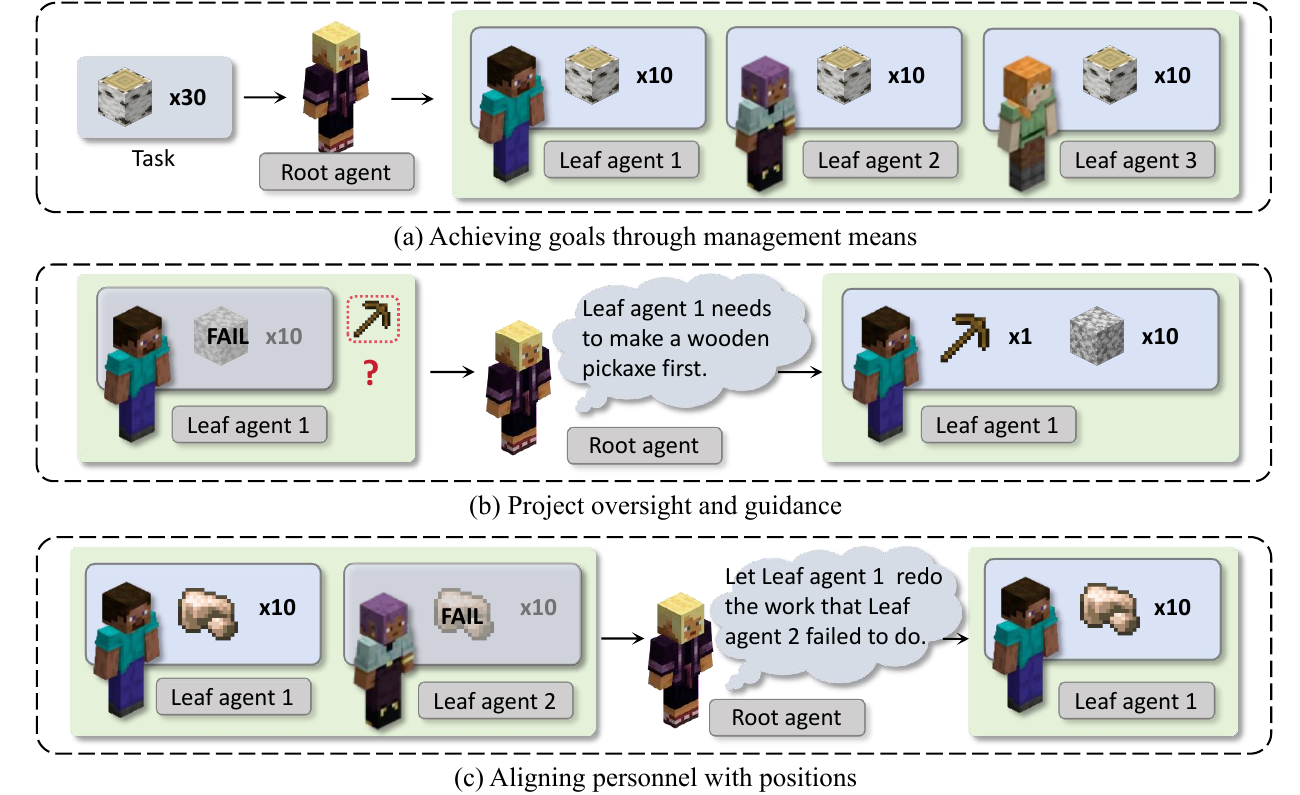}
	\end{center}
\vspace{-1em}
	\caption{
\textbf{Human-like leadership behaviors in self-organizing agent group.}}
 \vspace{-2em}
	\label{fig:behavior}
\end{figure*}
\subsection{Agent behaviors within organization}
\label{exp:behavious}

\subsubsection{Human-like leadership behaviors}
\label{exp:root_behavior}

\textbf{\textit{Achieving goals through management means:}}
In organizational management, leaders adopting a \textbf{\textit{non-hands-on approach}} to large-scale tasks is a rational consideration. As illustrated in Figure~\ref{fig:behavior}(a), the root agent in ToA, is tasked not only with solving individual issues but with equitably decomposing the workload. 
This property is achieved by instructing the task planning prompt of the leadership agent to assign tasks to leaf agents rather than itself. 
Refer to Appendix~\ref{sec:full_prompt_design} for details. 

\textbf{\textit{Project oversight and guidance:}}
As depicted in Figure~\ref{fig:behavior}(b), we observed that when the root agent learned of the leaf agent struggling to cope with a problem, the leader swiftly implemented a practical solution. 
%
%
This is mainly due to the progress monitor in the hourglass architecture, enabling the leader to clearly understand each leaf agent's status.

\textbf{\textit{Aligning personnel with positions:}}
Aligning suitable personnel with corresponding positions is a fundamental principle in human resources management.
In the context of an agent organization, as depicted in Figure~\ref{fig:behavior}(c), we observe a proactive leadership agent leveraging a nuanced understanding of past experiences and skills of team members to strategically allocate tasks.
\subsubsection{Behaving like human employees}
\label{exp:leaf_behavior}
During collaborative processes in agent organization, we've observed leaf agents displaying behaviors that mimic human employees. This is because, in ToA for the leaf agent, we require it to execute the tasks it receives autonomously. As a result, unlike agents in the GoA, it does not have a certain probability of delegating tasks to other agents. 
Moreover, regardless of whether the task execution is successful or not, the leaf agent proactively provides feedback, which aids the root agent in making further decisions. 
For more details, see Appendix~\ref{appendix:leaf_behavior}.
\section{Conclusion}
In conclusion, this study has introduced self-organizing agents (\model{}), a type of embodied agent group capable of autonomously orchestrating workflows without manual human design. 
The \model{} includes a tree-like organizational structure, an hourglass agent architecture, and a non-obstructive collaboration paradigm. 
Our experiments in the Minecraft environment have underscored its exceptional performance, showcasing superior capabilities across various tasks in comparison to individual agents and alternative organizational approaches. Importantly, we have observed anthropomorphic social behaviors in the organizational dynamics among the agents.
With the progression of artificial general intelligence, obtaining a deeper understanding of the organizational dynamics of scalable multi-agent systems becomes increasingly crucial. \model{} have commenced preliminary explorations in this field, and we foresee their potential adaptability and enhancement across a wider range of embodied tasks in the future.


\bibliography{cited}

\begin{thebibliography}{55}
\providecommand{\natexlab}[1]{#1}
\providecommand{\url}[1]{\texttt{#1}}
\expandafter\ifx\csname urlstyle\endcsname\relax
  \providecommand{\doi}[1]{doi: #1}\else
  \providecommand{\doi}{doi: \begingroup \urlstyle{rm}\Url}\fi

\bibitem[Akata et~al.(2023)Akata, Schulz, Coda-Forno, Oh, Bethge, and Schulz]{akata2023playing}
Elif Akata, Lion Schulz, Julian Coda-Forno, Seong~Joon Oh, Matthias Bethge, and Eric Schulz.
\newblock Playing repeated games with large language models.
\newblock \emph{arXiv preprint}, 2023.

\bibitem[Baker et~al.(2022)Baker, Akkaya, Zhokov, Huizinga, Tang, Ecoffet, Houghton, Sampedro, and Clune]{baker2022video}
Bowen Baker, Ilge Akkaya, Peter Zhokov, Joost Huizinga, Jie Tang, Adrien Ecoffet, Brandon Houghton, Raul Sampedro, and Jeff Clune.
\newblock Video pretraining (vpt): Learning to act by watching unlabeled online videos.
\newblock \emph{NeurIPS}, 2022.

\bibitem[Brohan et~al.(2023)Brohan, Brown, Carbajal, Chebotar, Chen, Choromanski, Ding, Driess, Dubey, Finn, et~al.]{brohan2023rt}
Anthony Brohan, Noah Brown, Justice Carbajal, Yevgen Chebotar, Xi~Chen, Krzysztof Choromanski, Tianli Ding, Danny Driess, Avinava Dubey, Chelsea Finn, et~al.
\newblock Rt-2: Vision-language-action models transfer web knowledge to robotic control.
\newblock \emph{arXiv preprint}, 2023.

\bibitem[Cai et~al.(2023)Cai, Wang, Ma, Chen, and Zhou]{cai2023large}
Tianle Cai, Xuezhi Wang, Tengyu Ma, Xinyun Chen, and Denny Zhou.
\newblock Large language models as tool makers.
\newblock \emph{arXiv preprint}, 2023.

\bibitem[Chen et~al.(2023{\natexlab{a}})Chen, Su, Zuo, Yang, Yuan, Qian, Chan, Qin, Lu, Xie, et~al.]{chen2023agentverse}
Weize Chen, Yusheng Su, Jingwei Zuo, Cheng Yang, Chenfei Yuan, Chen Qian, Chi-Min Chan, Yujia Qin, Yaxi Lu, Ruobing Xie, et~al.
\newblock Agentverse: Facilitating multi-agent collaboration and exploring emergent behaviors in agents.
\newblock \emph{arXiv preprint}, 2023{\natexlab{a}}.

\bibitem[Chen et~al.(2023{\natexlab{b}})Chen, Lin, Sch{\"a}rli, and Zhou]{chen2023teaching}
Xinyun Chen, Maxwell Lin, Nathanael Sch{\"a}rli, and Denny Zhou.
\newblock Teaching large language models to self-debug.
\newblock \emph{arXiv preprint}, 2023{\natexlab{b}}.

\bibitem[Chen et~al.(2023{\natexlab{c}})Chen, Arkin, Zhang, Roy, and Fan]{chen2023scalable}
Yongchao Chen, Jacob Arkin, Yang Zhang, Nicholas Roy, and Chuchu Fan.
\newblock Scalable multi-robot collaboration with large language models: Centralized or decentralized systems?
\newblock \emph{arXiv preprint}, 2023{\natexlab{c}}.

\bibitem[Chung et~al.(2022)Chung, Hou, Longpre, Zoph, Tay, Fedus, Li, Wang, Dehghani, Brahma, et~al.]{chung2022scaling}
Hyung~Won Chung, Le~Hou, Shayne Longpre, Barret Zoph, Yi~Tay, William Fedus, Yunxuan Li, Xuezhi Wang, Mostafa Dehghani, Siddhartha Brahma, et~al.
\newblock Scaling instruction-finetuned language models.
\newblock \emph{arXiv preprint}, 2022.

\bibitem[Dong et~al.(2023)Dong, Jiang, Jin, and Li]{dong2023self}
Yihong Dong, Xue Jiang, Zhi Jin, and Ge~Li.
\newblock Self-collaboration code generation via chatgpt.
\newblock \emph{arXiv preprint}, 2023.

\bibitem[Fan et~al.(2022)Fan, Wang, Jiang, Mandlekar, Yang, Zhu, Tang, Huang, Zhu, and Anandkumar]{fan2022minedojo}
Linxi Fan, Guanzhi Wang, Yunfan Jiang, Ajay Mandlekar, Yuncong Yang, Haoyi Zhu, Andrew Tang, De-An Huang, Yuke Zhu, and Anima Anandkumar.
\newblock Minedojo: Building open-ended embodied agents with internet-scale knowledge.
\newblock \emph{NeurIPS}, 2022.

\bibitem[Fehr \& G{\"a}chter(2000)Fehr and G{\"a}chter]{fehr2000cooperation}
Ernst Fehr and Simon G{\"a}chter.
\newblock Cooperation and punishment in public goods experiments.
\newblock \emph{American Economic Review}, 2000.

\bibitem[Fujita(2009)]{fujita2009intelligence}
Masahiro Fujita.
\newblock Intelligence dynamics: a concept and preliminary experiments for open-ended learning agents.
\newblock \emph{Autonomous Agents and Multi-Agent Systems}, 2009.

\bibitem[Gong et~al.(2023)Gong, Huang, Ma, Vo, Durante, Noda, Zheng, Zhu, Terzopoulos, Fei-Fei, et~al.]{gong2023mindagent}
Ran Gong, Qiuyuan Huang, Xiaojian Ma, Hoi Vo, Zane Durante, Yusuke Noda, Zilong Zheng, Song-Chun Zhu, Demetri Terzopoulos, Li~Fei-Fei, et~al.
\newblock Mindagent: Emergent gaming interaction.
\newblock \emph{arXiv preprint}, 2023.

\bibitem[Grossi et~al.(2006)Grossi, Dignum, Dignum, Dastani, and Royakkers]{grossi2006structural}
Davide Grossi, Frank Dignum, Virginia Dignum, Mehdi Dastani, and Lamb{\`e}r Royakkers.
\newblock Structural evaluation of agent organizations.
\newblock In \emph{IJCAI}, 2006.

\bibitem[Hong et~al.(2023)Hong, Zheng, Chen, Cheng, Zhang, Wang, Yau, Lin, Zhou, Ran, et~al.]{hong2023metagpt}
Sirui Hong, Xiawu Zheng, Jonathan Chen, Yuheng Cheng, Ceyao Zhang, Zili Wang, Steven Ka~Shing Yau, Zijuan Lin, Liyang Zhou, Chenyu Ran, et~al.
\newblock Metagpt: Meta programming for multi-agent collaborative framework.
\newblock \emph{arXiv preprint}, 2023.

\bibitem[Jensen et~al.(2017)Jensen, Dignum, and Villadsen]{jensen2017framework}
Andreas~Schmidt Jensen, Virginia Dignum, and J{\o}rgen Villadsen.
\newblock A framework for organization-aware agents.
\newblock \emph{Autonomous Agents and Multi-Agent Systems}, 2017.

\bibitem[Ji et~al.(2023)Ji, Zhuge, Gao, Fan, Sakaridis, and Gool]{ji2023masked}
Ge-Peng Ji, Mingchen Zhuge, Dehong Gao, Deng-Ping Fan, Christos Sakaridis, and Luc~Van Gool.
\newblock Masked vision-language transformer in fashion.
\newblock \emph{Machine Intelligence Research}, 2023.

\bibitem[Johnson et~al.(2016)Johnson, Hofmann, Hutton, and Bignell]{johnson2016malmo}
Matthew Johnson, Katja Hofmann, Tim Hutton, and David Bignell.
\newblock The malmo platform for artificial intelligence experimentation.
\newblock In \emph{IJCAI}, 2016.

\bibitem[Kanervisto et~al.(2022)Kanervisto, Milani, Ramanauskas, Topin, Lin, Li, Shi, Ye, Fu, Yang, et~al.]{kanervisto2022minerl}
Anssi Kanervisto, Stephanie Milani, Karolis Ramanauskas, Nicholay Topin, Zichuan Lin, Junyou Li, Jianing Shi, Deheng Ye, Qiang Fu, Wei Yang, et~al.
\newblock Minerl diamond 2021 competition: Overview, results, and lessons learned.
\newblock \emph{NeurIPS 2021 Competitions and Demonstrations Track}, 2022.

\bibitem[Kanitscheider et~al.(2021)Kanitscheider, Huizinga, Farhi, Guss, Houghton, Sampedro, Zhokhov, Baker, Ecoffet, Tang, et~al.]{kanitscheider2021multi}
Ingmar Kanitscheider, Joost Huizinga, David Farhi, William~Hebgen Guss, Brandon Houghton, Raul Sampedro, Peter Zhokhov, Bowen Baker, Adrien Ecoffet, Jie Tang, et~al.
\newblock Multi-task curriculum learning in a complex, visual, hard-exploration domain: Minecraft.
\newblock \emph{arXiv preprint}, 2021.

\bibitem[Kasneci et~al.(2023)Kasneci, Se{\ss}ler, K{\"u}chemann, Bannert, Dementieva, Fischer, Gasser, Groh, G{\"u}nnemann, H{\"u}llermeier, et~al.]{kasneci2023chatgpt}
Enkelejda Kasneci, Kathrin Se{\ss}ler, Stefan K{\"u}chemann, Maria Bannert, Daryna Dementieva, Frank Fischer, Urs Gasser, Georg Groh, Stephan G{\"u}nnemann, Eyke H{\"u}llermeier, et~al.
\newblock Chatgpt for good? on opportunities and challenges of large language models for education.
\newblock \emph{Learning and individual differences}, 2023.

\bibitem[Kiseleva et~al.(2022{\natexlab{a}})Kiseleva, Li, Aliannejadi, Mohanty, ter Hoeve, Burtsev, Skrynnik, Zholus, Panov, Srinet, et~al.]{kiseleva2022interactive}
Julia Kiseleva, Ziming Li, Mohammad Aliannejadi, Shrestha Mohanty, Maartje ter Hoeve, Mikhail Burtsev, Alexey Skrynnik, Artem Zholus, Aleksandr Panov, Kavya Srinet, et~al.
\newblock Interactive grounded language understanding in a collaborative environment: Iglu 2021.
\newblock In \emph{NeurIPS 2021 Competitions and Demonstrations Track}, 2022{\natexlab{a}}.

\bibitem[Kiseleva et~al.(2022{\natexlab{b}})Kiseleva, Skrynnik, Zholus, Mohanty, Arabzadeh, Côté, Aliannejadi, Teruel, Li, Burtsev, ter Hoeve, Volovikova, Panov, Sun, Srinet, Szlam, and Awadallah]{kiseleva2022iglu}
Julia Kiseleva, Alexey Skrynnik, Artem Zholus, Shrestha Mohanty, Negar Arabzadeh, Marc-Alexandre Côté, Mohammad Aliannejadi, Milagro Teruel, Ziming Li, Mikhail Burtsev, Maartje ter Hoeve, Zoya Volovikova, Aleksandr Panov, Yuxuan Sun, Kavya Srinet, Arthur Szlam, and Ahmed Awadallah.
\newblock Iglu 2022: Interactive grounded language understanding in a collaborative environment at neurips 2022, 2022{\natexlab{b}}.

\bibitem[Li et~al.(2023)Li, Hammoud, Itani, Khizbullin, and Ghanem]{li2023camel}
Guohao Li, Hasan Abed Al~Kader Hammoud, Hani Itani, Dmitrii Khizbullin, and Bernard Ghanem.
\newblock Camel: Communicative agents for" mind" exploration of large scale language model society.
\newblock \emph{arXiv preprint}, 2023.

\bibitem[Lin et~al.(2021)Lin, Li, Shi, Ye, Fu, and Yang]{lin2021juewu}
Zichuan Lin, Junyou Li, Jianing Shi, Deheng Ye, Qiang Fu, and Wei Yang.
\newblock Juewu-mc: Playing minecraft with sample-efficient hierarchical reinforcement learning.
\newblock \emph{arXiv preprint}, 2021.

\bibitem[Lu et~al.(2023)Lu, Wang, Liu, Lu, and Tang]{lu2023thinkbot}
Guanxing Lu, Ziwei Wang, Changliu Liu, Jiwen Lu, and Yansong Tang.
\newblock Thinkbot: Embodied instruction following with thought chain reasoning.
\newblock \emph{arXiv preprint}, 2023.

\bibitem[Mandi et~al.(2023)Mandi, Jain, and Song]{mandi2023roco}
Zhao Mandi, Shreeya Jain, and Shuran Song.
\newblock Roco: Dialectic multi-robot collaboration with large language models.
\newblock \emph{arXiv preprint}, 2023.

\bibitem[Mao et~al.(2022)Mao, Wang, Hao, Mao, Lu, Wu, Hao, Li, and Tang]{mao2022seihai}
Hangyu Mao, Chao Wang, Xiaotian Hao, Yihuan Mao, Yiming Lu, Chengjie Wu, Jianye Hao, Dong Li, and Pingzhong Tang.
\newblock Seihai: A sample-efficient hierarchical ai for the minerl competition.
\newblock In \emph{DAI}, 2022.

\bibitem[Mohanty et~al.(2022)Mohanty, Arabzadeh, Teruel, Sun, Zholus, Skrynnik, Burtsev, Srinet, Panov, Szlam, et~al.]{mohanty2022collecting}
Shrestha Mohanty, Negar Arabzadeh, Milagro Teruel, Yuxuan Sun, Artem Zholus, Alexey Skrynnik, Mikhail Burtsev, Kavya Srinet, Aleksandr Panov, Arthur Szlam, et~al.
\newblock Collecting interactive multi-modal datasets for grounded language understanding.
\newblock \emph{arXiv preprint}, 2022.

\bibitem[Mohanty et~al.(2023)Mohanty, Arabzadeh, Kiseleva, Zholus, Teruel, Awadallah, Sun, Srinet, and Szlam]{mohanty2023transforming}
Shrestha Mohanty, Negar Arabzadeh, Julia Kiseleva, Artem Zholus, Milagro Teruel, Ahmed Awadallah, Yuxuan Sun, Kavya Srinet, and Arthur Szlam.
\newblock Transforming human-centered ai collaboration: Redefining embodied agents capabilities through interactive grounded language instructions.
\newblock \emph{arXiv preprint}, 2023.

\bibitem[OpenAI(2023{\natexlab{a}})]{introduce2022chatgpt}
OpenAI.
\newblock Introducing chatgpt, 2023{\natexlab{a}}.

\bibitem[OpenAI(2023{\natexlab{b}})]{openai2023gpt4}
OpenAI.
\newblock Gpt-4 technical report, 2023{\natexlab{b}}.

\bibitem[Ouyang et~al.(2022)Ouyang, Wu, Jiang, Almeida, Wainwright, Mishkin, Zhang, Agarwal, Slama, Ray, et~al.]{ouyang2022training}
Long Ouyang, Jeffrey Wu, Xu~Jiang, Diogo Almeida, Carroll Wainwright, Pamela Mishkin, Chong Zhang, Sandhini Agarwal, Katarina Slama, Alex Ray, et~al.
\newblock Training language models to follow instructions with human feedback.
\newblock \emph{NeurIPS}, 2022.

\bibitem[Padmakumar et~al.(2023)Padmakumar, Inan, Gella, Lange, and Hakkani-Tur]{padmakumar2023multimodal}
Aishwarya Padmakumar, Mert Inan, Spandana Gella, Patrick~L Lange, and Dilek Hakkani-Tur.
\newblock Multimodal embodied plan prediction augmented with synthetic embodied dialogue.
\newblock In \emph{Conference on Empirical Methods in Natural Language Processing}, 2023.

\bibitem[Park et~al.(2023)Park, O'Brien, Cai, Morris, Liang, and Bernstein]{park2023generative}
Joon~Sung Park, Joseph~C. O'Brien, Carrie~J. Cai, Meredith~Ringel Morris, Percy Liang, and Michael~S. Bernstein.
\newblock Generative agents: Interactive simulacra of human behavior, 2023.

\bibitem[Perez-Liebana et~al.(2019)Perez-Liebana, Hofmann, Mohanty, Kuno, Kramer, Devlin, Gaina, and Ionita]{perez2019multi}
Diego Perez-Liebana, Katja Hofmann, Sharada~Prasanna Mohanty, Noburu Kuno, Andre Kramer, Sam Devlin, Raluca~D Gaina, and Daniel Ionita.
\newblock The multi-agent reinforcement learning in malm$\backslash$" o (marl$\backslash$" o) competition.
\newblock \emph{arXiv preprint}, 2019.

\bibitem[PrismarineJS(2013)]{PrismarineJS}
PrismarineJS.
\newblock Prismarinejs. prismarinejs/mineflayer: Create minecraft bots with a powerful, stable, and high level javascript api, 2013.

\bibitem[Qian et~al.(2023)Qian, Cong, Yang, Chen, Su, Xu, Liu, and Sun]{qian2023communicative}
Chen Qian, Xin Cong, Cheng Yang, Weize Chen, Yusheng Su, Juyuan Xu, Zhiyuan Liu, and Maosong Sun.
\newblock Communicative agents for software development.
\newblock \emph{arXiv preprint}, 2023.

\bibitem[Shinn et~al.(2023)Shinn, Labash, and Gopinath]{shinn2023reflexion}
Noah Shinn, Beck Labash, and Ashwin Gopinath.
\newblock Reflexion: an autonomous agent with dynamic memory and self-reflection.
\newblock \emph{arXiv preprint}, 2023.

\bibitem[Tang et~al.(2023)Tang, Zou, Zhang, Zhao, Zhang, Cohan, and Gerstein]{tang2023medagents}
Xiangru Tang, Anni Zou, Zhuosheng Zhang, Yilun Zhao, Xingyao Zhang, Arman Cohan, and Mark Gerstein.
\newblock Medagents: Large language models as collaborators for zero-shot medical reasoning.
\newblock \emph{arXiv preprint}, 2023.

\bibitem[Touvron et~al.(2023)Touvron, Lavril, Izacard, Martinet, Lachaux, Lacroix, Rozi{\`e}re, Goyal, Hambro, Azhar, et~al.]{touvron2023llama}
Hugo Touvron, Thibaut Lavril, Gautier Izacard, Xavier Martinet, Marie-Anne Lachaux, Timoth{\'e}e Lacroix, Baptiste Rozi{\`e}re, Naman Goyal, Eric Hambro, Faisal Azhar, et~al.
\newblock Llama: Open and efficient foundation language models.
\newblock \emph{arXiv preprint}, 2023.

\bibitem[Wang et~al.(2023{\natexlab{a}})Wang, Xie, Jiang, Mandlekar, Xiao, Zhu, Fan, and Anandkumar]{wang2023voyager}
Guanzhi Wang, Yuqi Xie, Yunfan Jiang, Ajay Mandlekar, Chaowei Xiao, Yuke Zhu, Linxi Fan, and Anima Anandkumar.
\newblock Voyager: An open-ended embodied agent with large language models.
\newblock \emph{arXiv preprint}, 2023{\natexlab{a}}.

\bibitem[Wang et~al.(2023{\natexlab{b}})Wang, Cai, Liu, Jin, Hou, Zhang, Lin, He, Zheng, Yang, Ma, and Liang]{wang2023jarvis1}
Zihao Wang, Shaofei Cai, Anji Liu, Yonggang Jin, Jinbing Hou, Bowei Zhang, Haowei Lin, Zhaofeng He, Zilong Zheng, Yaodong Yang, Xiaojian Ma, and Yitao Liang.
\newblock Jarvis-1: Open-world multi-task agents with memory-augmented multimodal language models.
\newblock \emph{arXiv preprint}, 2023{\natexlab{b}}.

\bibitem[Wang et~al.(2023{\natexlab{c}})Wang, Cai, Liu, Ma, and Liang]{wang2023describe}
Zihao Wang, Shaofei Cai, Anji Liu, Xiaojian Ma, and Yitao Liang.
\newblock Describe, explain, plan and select: Interactive planning with large language models enables open-world multi-task agents.
\newblock \emph{arXiv preprint}, 2023{\natexlab{c}}.

\bibitem[Wei et~al.(2022)Wei, Wang, Schuurmans, Bosma, Xia, Chi, Le, Zhou, et~al.]{wei2022chain}
Jason Wei, Xuezhi Wang, Dale Schuurmans, Maarten Bosma, Fei Xia, Ed~Chi, Quoc~V Le, Denny Zhou, et~al.
\newblock Chain-of-thought prompting elicits reasoning in large language models.
\newblock \emph{NeurIPS}, 2022.

\bibitem[Weinbaum \& Veitas(2017)Weinbaum and Veitas]{weinbaum2017open}
David Weinbaum and Viktoras Veitas.
\newblock Open ended intelligence: the individuation of intelligent agents.
\newblock \emph{Journal of Experimental \& Theoretical Artificial Intelligence}, 2017.

\bibitem[Woolley et~al.(2010)Woolley, Chabris, Pentland, Hashmi, and Malone]{woolley2010evidence}
Anita~Williams Woolley, Christopher~F Chabris, Alex Pentland, Nada Hashmi, and Thomas~W Malone.
\newblock Evidence for a collective intelligence factor in the performance of human groups.
\newblock \emph{science}, 2010.

\bibitem[Wu et~al.(2023)Wu, Bansal, Zhang, Wu, Zhang, Zhu, Li, Jiang, Zhang, and Wang]{wu2023autogen}
Qingyun Wu, Gagan Bansal, Jieyu Zhang, Yiran Wu, Shaokun Zhang, Erkang Zhu, Beibin Li, Li~Jiang, Xiaoyun Zhang, and Chi Wang.
\newblock Autogen: Enabling next-gen llm applications via multi-agent conversation framework.
\newblock \emph{arXiv preprint}, 2023.

\bibitem[Xiang et~al.(2023)Xiang, Tao, Gu, Shu, Wang, Yang, and Hu]{xiang2023language}
Jiannan Xiang, Tianhua Tao, Yi~Gu, Tianmin Shu, Zirui Wang, Zichao Yang, and Zhiting Hu.
\newblock Language models meet world models: Embodied experiences enhance language models.
\newblock \emph{arXiv preprint}, 2023.

\bibitem[Yao et~al.(2023)Yao, Yu, Zhao, Shafran, Griffiths, Cao, and Narasimhan]{yao2023tree}
Shunyu Yao, Dian Yu, Jeffrey Zhao, Izhak Shafran, Thomas~L Griffiths, Yuan Cao, and Karthik Narasimhan.
\newblock Tree of thoughts: Deliberate problem solving with large language models.
\newblock \emph{arXiv preprint}, 2023.

\bibitem[Zeng et~al.(2022)Zeng, Attarian, Ichter, Choromanski, Wong, Welker, Tombari, Purohit, Ryoo, Sindhwani, et~al.]{zeng2022socratic}
Andy Zeng, Maria Attarian, Brian Ichter, Krzysztof Choromanski, Adrian Wong, Stefan Welker, Federico Tombari, Aveek Purohit, Michael Ryoo, Vikas Sindhwani, et~al.
\newblock Socratic models: Composing zero-shot multimodal reasoning with language.
\newblock \emph{arXiv preprint}, 2022.

\bibitem[Zhang et~al.(2023)Zhang, Du, Shan, Zhou, Du, Tenenbaum, Shu, and Gan]{zhang2023building}
Hongxin Zhang, Weihua Du, Jiaming Shan, Qinhong Zhou, Yilun Du, Joshua~B Tenenbaum, Tianmin Shu, and Chuang Gan.
\newblock Building cooperative embodied agents modularly with large language models.
\newblock \emph{arXiv preprint}, 2023.

\bibitem[Zhao et~al.(2023)Zhao, Chai, Wang, Boyi, Hao, Cao, Ye, Hwang, and Wang]{zhao2023see}
Zhonghan Zhao, Wenhao Chai, Xuan Wang, Li~Boyi, Shengyu Hao, Shidong Cao, Tian Ye, Jenq-Neng Hwang, and Gaoang Wang.
\newblock See and think: Embodied agent in virtual environment.
\newblock \emph{arXiv preprint}, 2023.

\bibitem[Zhu et~al.(2023)Zhu, Chen, Tian, Tao, Su, Yang, Huang, Li, Lu, Wang, et~al.]{zhu2023ghost}
Xizhou Zhu, Yuntao Chen, Hao Tian, Chenxin Tao, Weijie Su, Chenyu Yang, Gao Huang, Bin Li, Lewei Lu, Xiaogang Wang, et~al.
\newblock Ghost in the minecraft: Generally capable agents for open-world enviroments via large language models with text-based knowledge and memory.
\newblock \emph{arXiv preprint}, 2023.

\bibitem[Zhuge et~al.(2023)Zhuge, Liu, Faccio, Ashley, Csord{\'a}s, Gopalakrishnan, Hamdi, Hammoud, Herrmann, Irie, et~al.]{zhuge2023mindstorms}
Mingchen Zhuge, Haozhe Liu, Francesco Faccio, Dylan~R Ashley, R{\'o}bert Csord{\'a}s, Anand Gopalakrishnan, Abdullah Hamdi, Hasan Abed Al~Kader Hammoud, Vincent Herrmann, Kazuki Irie, et~al.
\newblock Mindstorms in natural language-based societies of mind.
\newblock \emph{arXiv preprint}, 2023.

\end{thebibliography}
\bibliographystyle{iclr2024_conference}

\newpage
\appendix
\label{sec:appendix}
\section{Example}
\subsection{Example of planning process}
\label{appendix:planning_process}
\begin{figure*}[h]
	\begin{center}
		\includegraphics[width=1.0\linewidth]{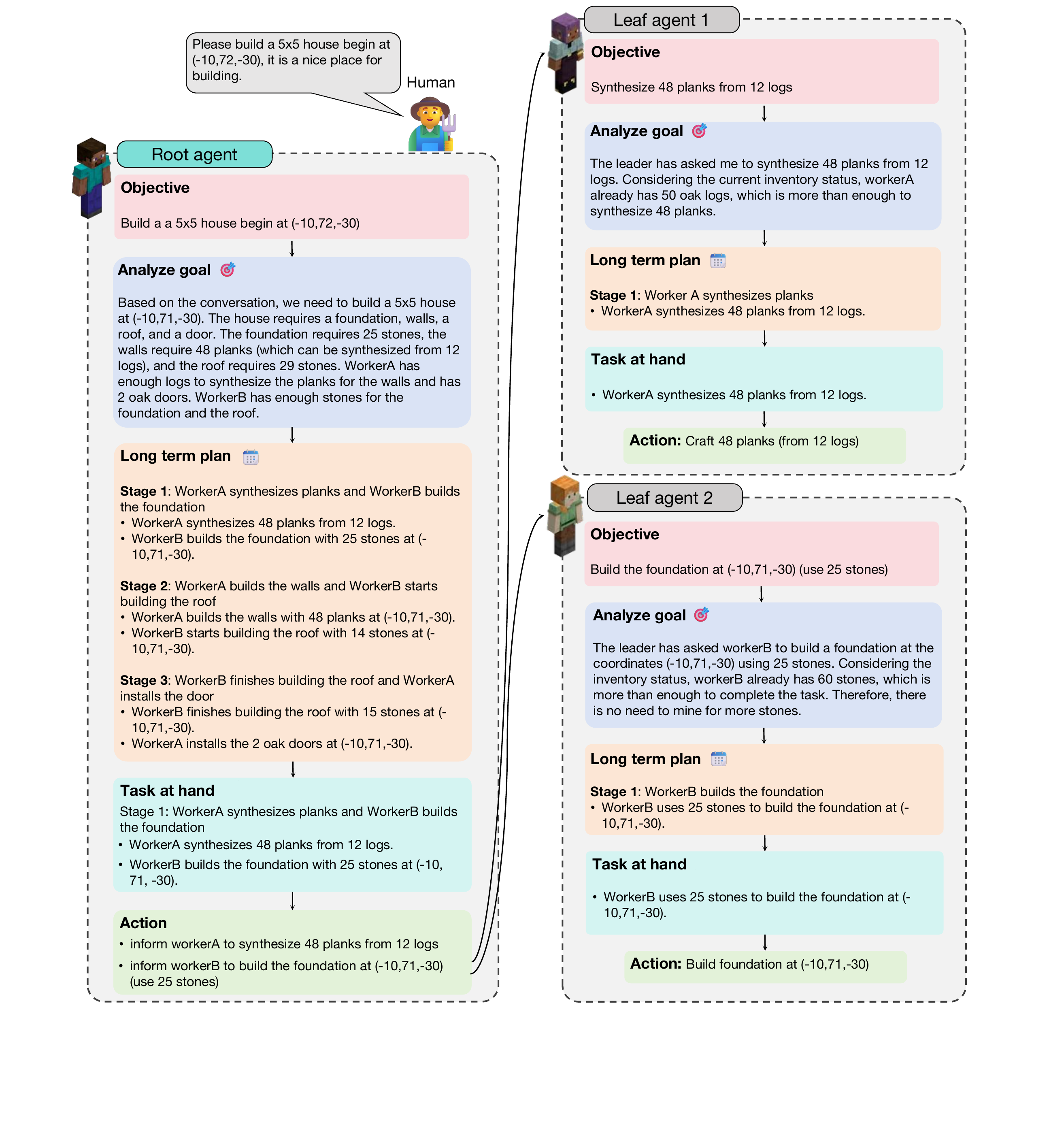}
	\end{center}
\vspace{-1em}
	\caption{
\textbf{The illustration of an example process of collaborative house building.} Full prompts are presented in Appendix~\ref{sec:full_prompt_design}.}
 \vspace{-1em}
	\label{fig:detail}
\end{figure*}

Figure~\ref{fig:detail} demonstrates the planning process of both the root agent and leaf agents in the ``build a house" task.

\subsection{Example of execution process}
\label{appendix:execution_process}
\begin{figure*}[htb]
	\def \imwidth {4.6cm}
	\def \imheight {2.3cm}
	\centering
	\includegraphics[height=\imheight, width=\imwidth]{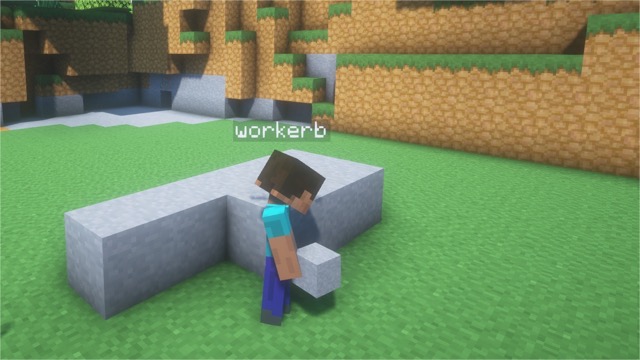}
	\includegraphics[height=\imheight, width=\imwidth]{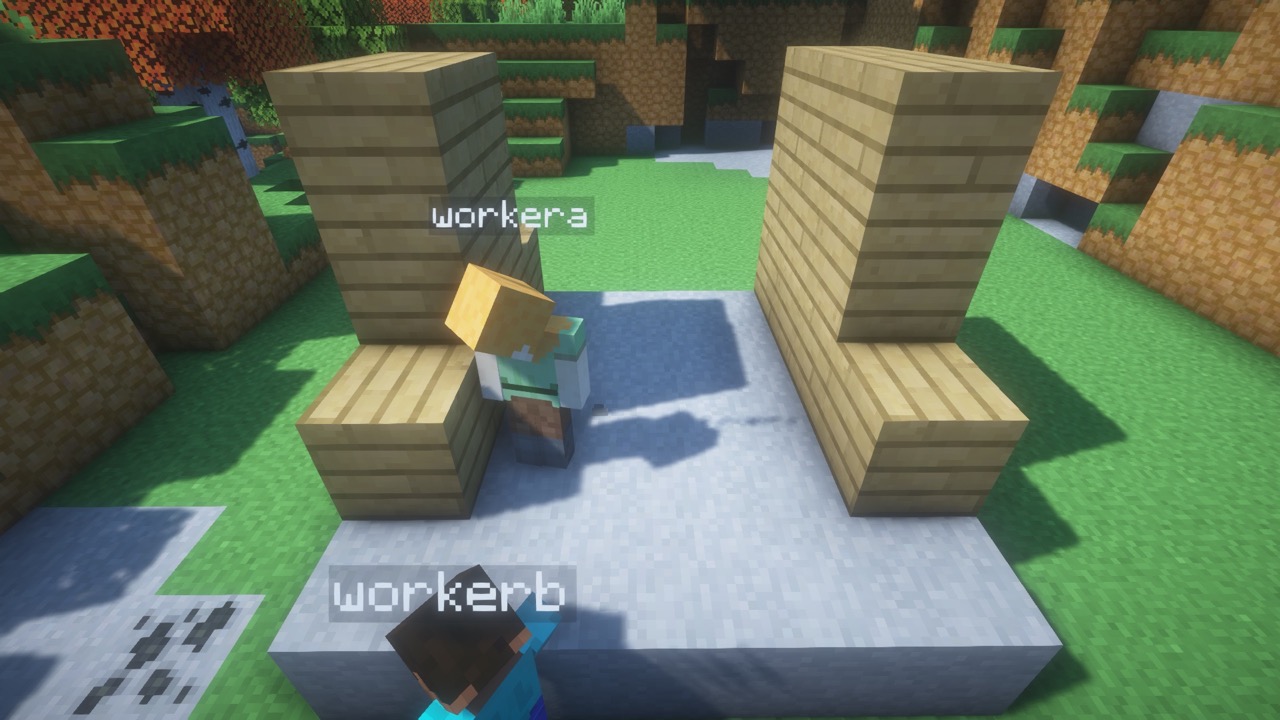}
	\includegraphics[height=\imheight, width=\imwidth]{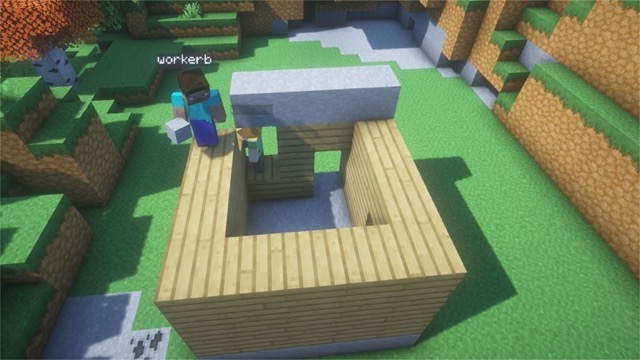}
	 \\
	\rotatebox{0}{\textcolor{white}{}Build foundation}
	\rotatebox{0}{\textcolor{white}{----------------------}Build walls}
	\rotatebox{0}{\textcolor{white}{-------------------------}Build roof}
	
	\caption{\textbf{The procedure of building a house.} The root agent commands while the two leaf agents collaborate to complete the construction.
	}
	\label{fig:building_procedure}
\end{figure*}

\begin{figure*}[htb]
	\def \imwidth {4.6cm}
	\def \imheight {2.3cm}
	\centering
	\includegraphics[height=\imheight, width=\imwidth]{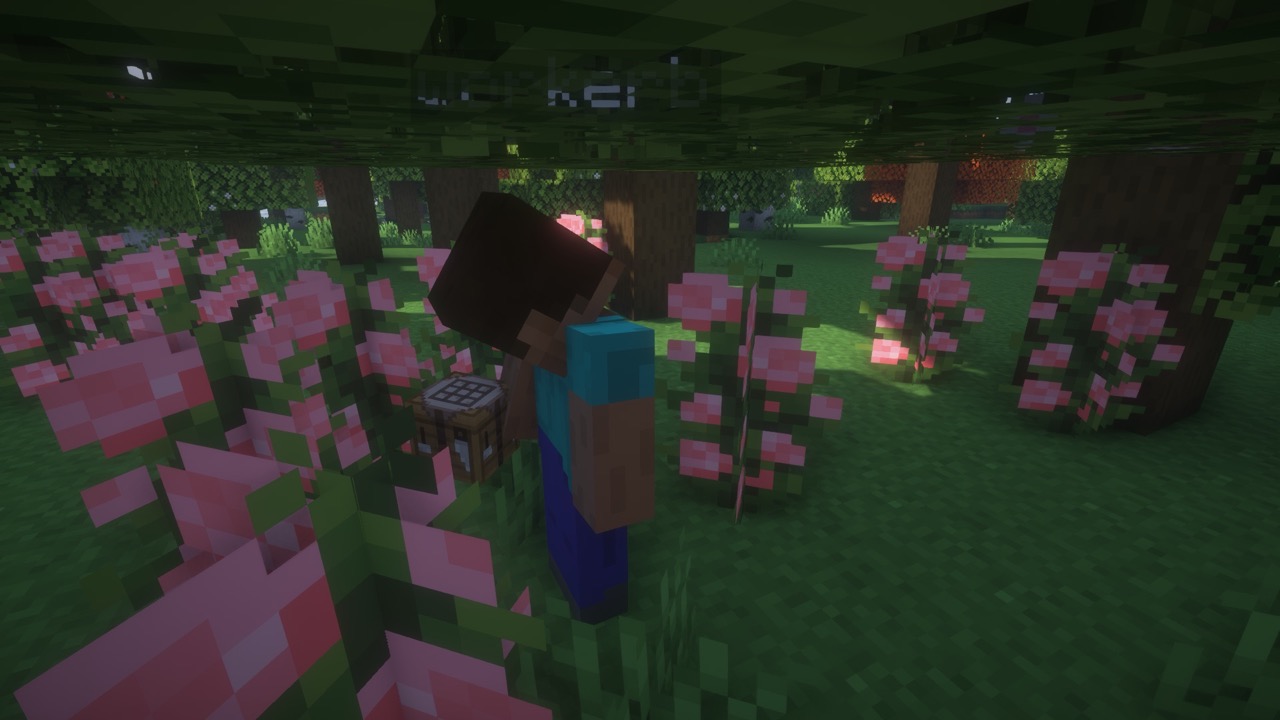}
	\includegraphics[height=\imheight, width=\imwidth]{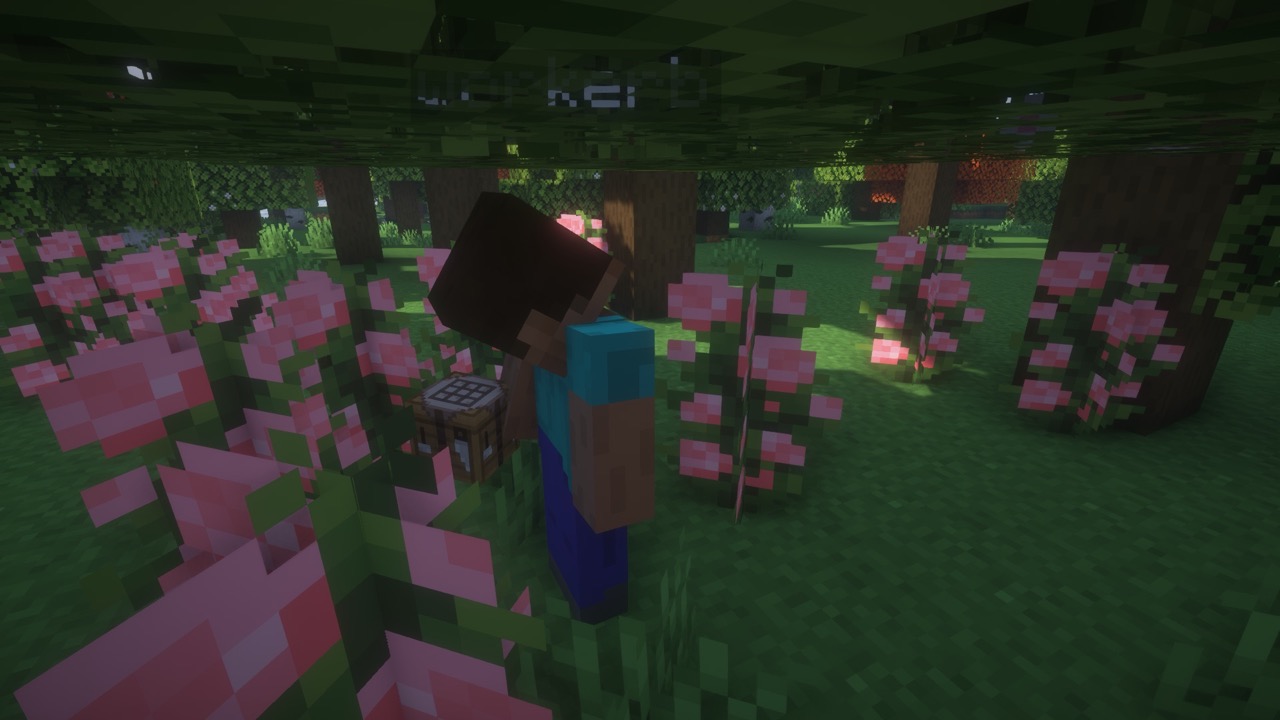}
 \includegraphics[height=\imheight, width=\imwidth]{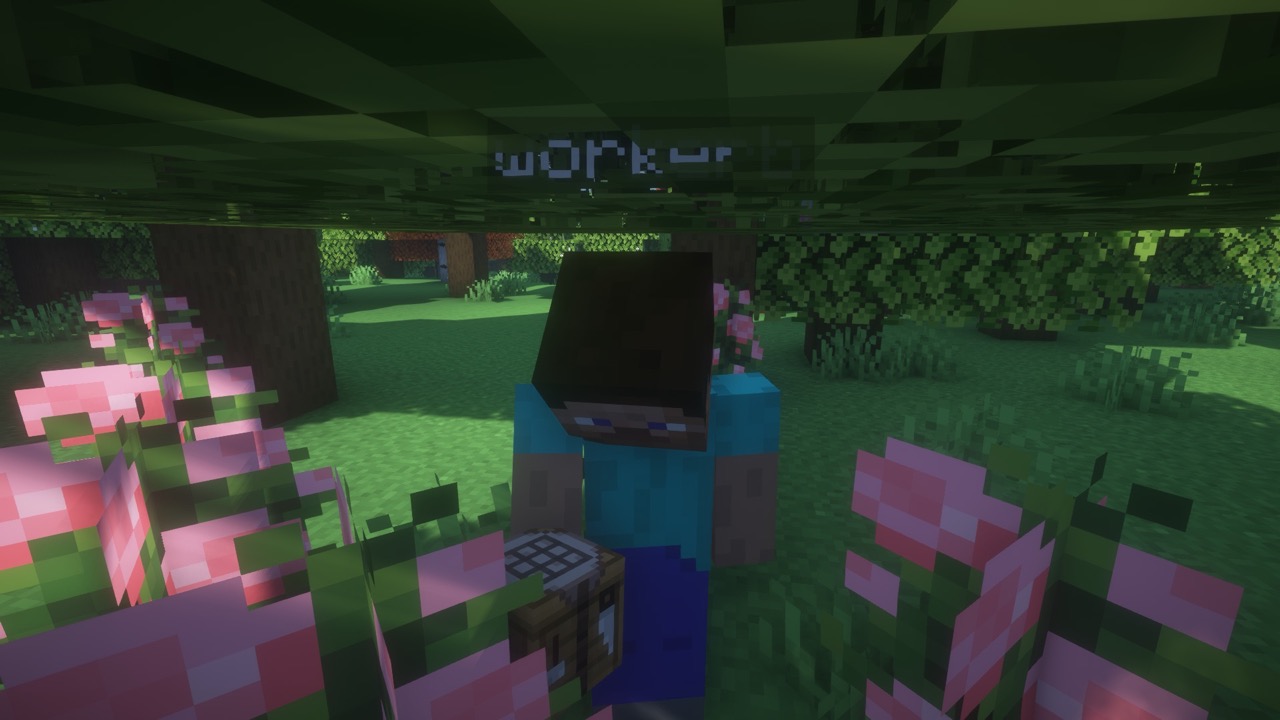}\\
 \rotatebox{0}{\textcolor{white}{---------}Mine logs}
	\rotatebox{0}{\textcolor{white}{-----------------------}Craft crafting table}
	\rotatebox{0}{\textcolor{white}{---------------}Craft wooden pickaxe}\\
	\includegraphics[height=\imheight, width=\imwidth]{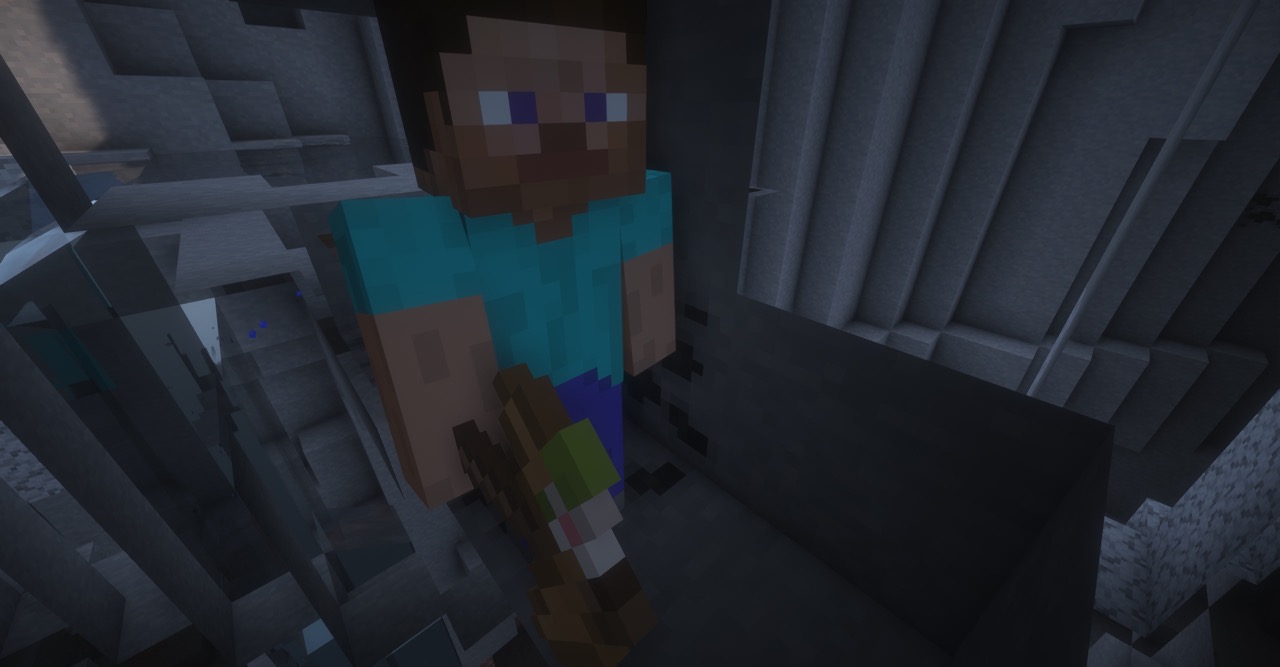}
 \includegraphics[height=\imheight, width=\imwidth]{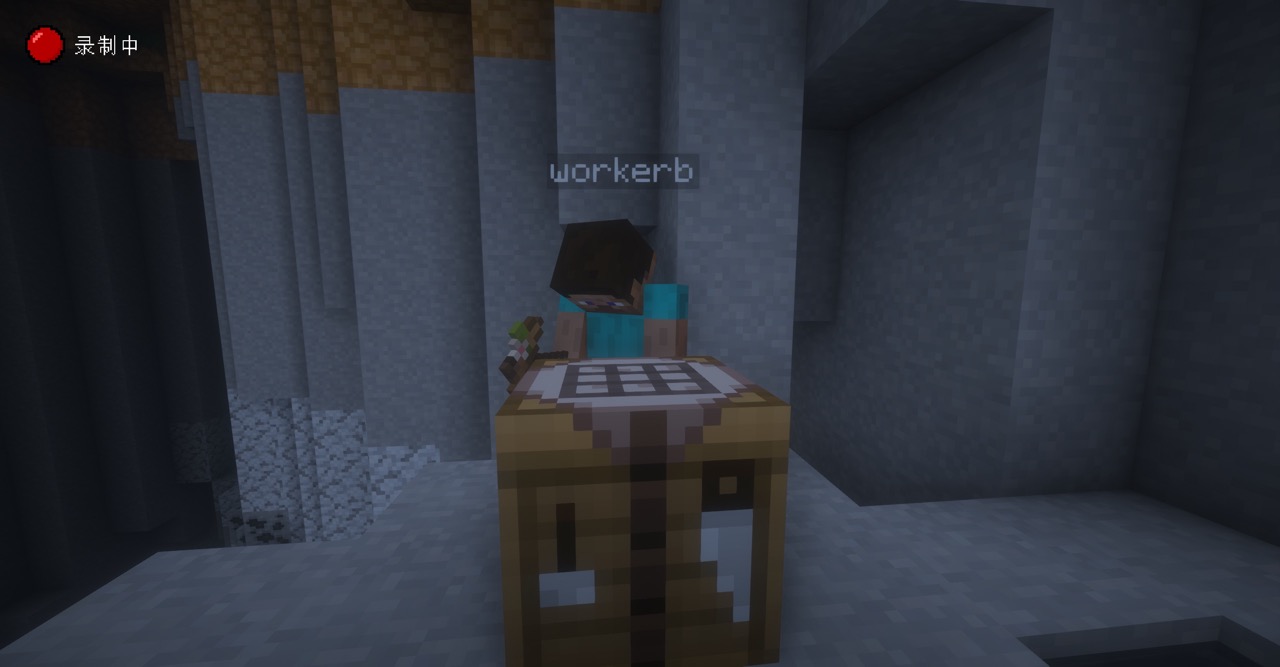}
 \includegraphics[height=\imheight, width=\imwidth]{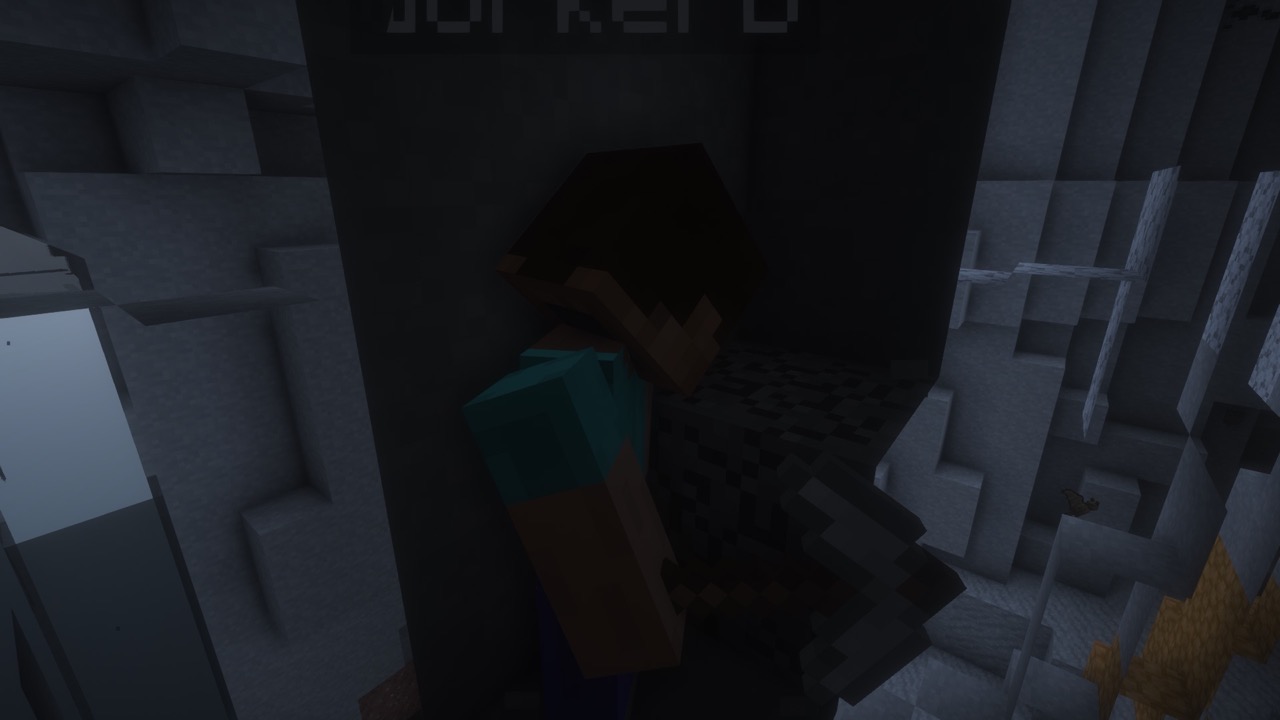}
    \rotatebox{0}{\textcolor{white}{}Mine stones}
	\rotatebox{0}{\textcolor{white}{--------------------}Craft stone pickaxe}
	\rotatebox{0}{\textcolor{white}{----------------------}Mine iron}

	\caption{\textbf{The procedure of mining iron.} As the execution processes of the two agents are similar, only one of them is presented here.
	}
	\label{fig:iron_procedure}
\end{figure*}

Figure~\ref{fig:building_procedure} and Figure~\ref{fig:iron_procedure} illustrate the collaborative processes of building a house and collecting iron, respectively.




\subsection{Agent behavior mimicking human employees}
\label{appendix:leaf_behavior}
\begin{figure*}[t]
	\begin{center}
		\includegraphics[width=1.0\linewidth]{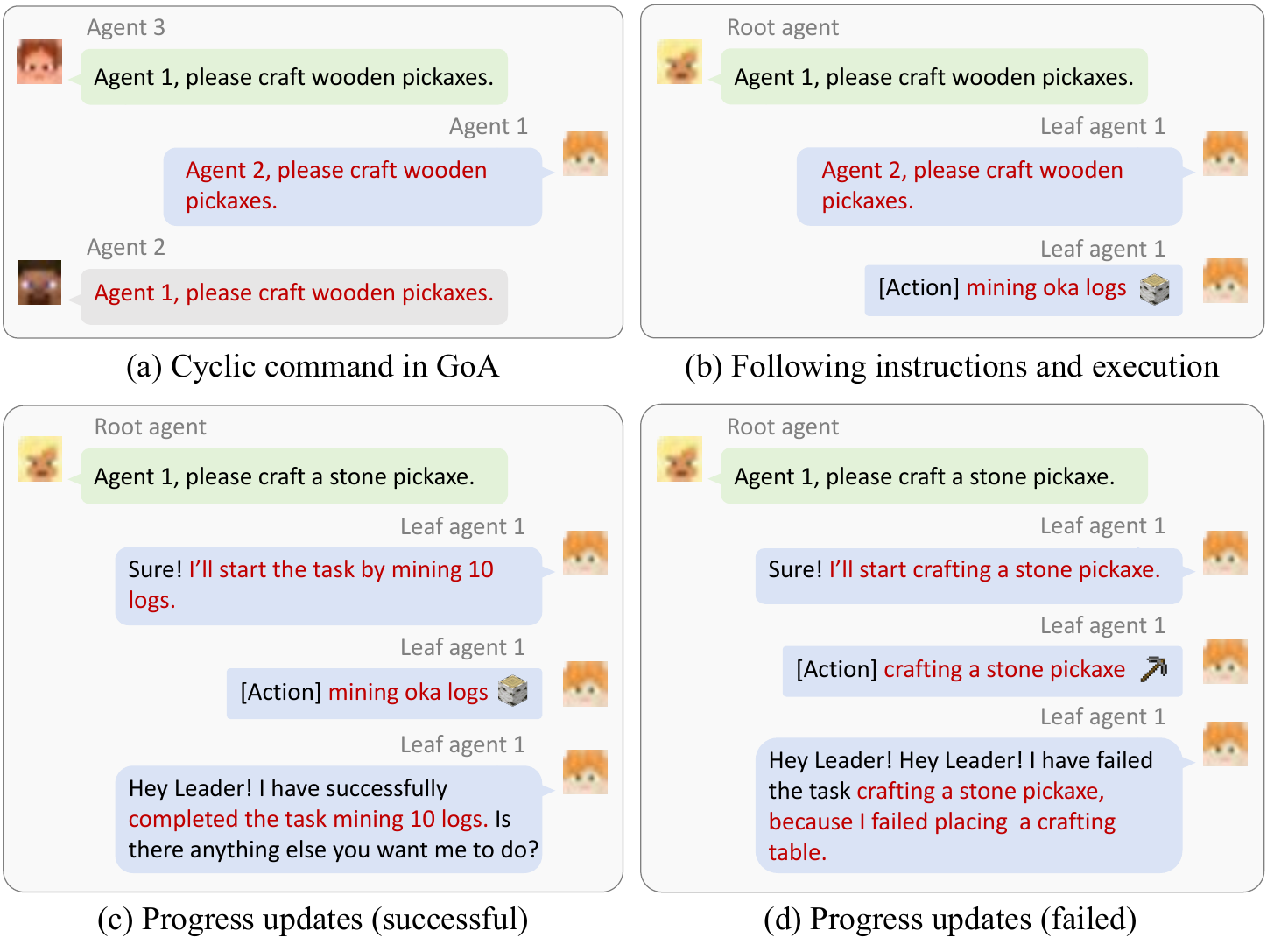}
	\end{center}
\vspace{-1em}
	\caption{
\textbf{Agent behaviors that mimic human employees.}}
 \vspace{-1em}
	\label{fig:example}
\end{figure*}
The following is a detailed description of behaviors similar to those of human employees exhibited by the LLM agent during task execution.

\textbf{\textit{Following instructions and execution:}}
In the human workplace, proactively following the instructions of a leader and possessing excellent execution capabilities ensure the achievement of organizational objectives.  
Within a GoA, as depicted in Figure~\ref{fig:example}(a), when leaf agent $a_{l1}$ receives instructions to craft a wooden pickaxe, he delegates the task to leaf agent $a_{l2}$. 
Subsequently, leaf agent $a_{l2}$ replicates this behavior, resulting in a loop of delegation, leading to inaction and production halt.
In contrast, as shown in Figure~\ref{fig:example}(b), in a ToA, when leaf agent $a_{l1}$ is assigned the task of crafting a wooden pickaxe, he promptly begins the task himself, following the directive of leadership agent without further delegation. 

\textbf{\textit{Progress updates to the leader:}}
The root agent needs to model the execution progress on time, and correspondingly, leaf agents are required to report the progress of task execution to the Root agent. 
As illustrated in Figure~\ref{fig:example}(c,d), a leaf agent reports at the commencement of a task and provides updates upon its completion, indicating success or failure. 
In case of task failure, the leaf agent also elucidates the reasons and current state of their inventory.

\section{Full prompt design}
\label{sec:full_prompt_design}
The displayed prompts have been corrected for minor grammar errors.
\subsection{Progress monitor}
\subsubsection{Components in the Prompt}
The input prompt to LLM consists of several components:
\begin{enumerate}[label=(\arabic*)]
\item Task to be inquired, proposed by the task planner.
\item Recent conversation, conversation since last task planning. Leaf agents report their task progress and current inventory status during conversations, allowing for the inference of task completion from the dialogue.
\item Chest information. In rare instances, leaf agents might place the results of their actions in nearby chests, necessitating the integration of chest information for accurate task completion assessment. In most cases, this field is left empty.
\end{enumerate}

\subsubsection{Full Prompt}
\label{full_prompt}
\begin{tcolorbox}[title={\textbf{\small Template of progress monitor}}, colback=whitesmoke, colframe=bluegray, boxrule=2pt, arc=0mm, breakable]
\scriptsize
\begin{lstlisting}[style=mystyle]
You're a judge of progress in Minecraft, and you're adept at judging whether or not [Task to be inquired] is complete based on the [Conversation] and [chest information] so far.

Your task is to perform the following actions to help me play Minecraft:
I'll give you [Task to be inquired], [Conversation] with another player, and the [Chest information] nearby.
First, please give a [Task result judgment] about the current progress of [Task to be inquired].
You should give your judgment right on the spot, not beat around the bush!
Then, determine the [Final task status] ('success' or 'fail' or 'unknown'), based on the progress analysis.

You must follow the following criteria:
1. Please focus only on [Conversation] or [Chest information] related to the [Task to be inquired], other information is not included in the analysis.
2. As long as one of the [Conversation] and [Chest information] contains valid information about [Task to be inquired], it is sufficient for [Progress analysis] and [Final task status].
3. Recieve the response of 'I will start task: xxx' in [Conversation] means that the task has started, [Final task status] should be unknown.
4. No response or only get 'Got it!', means that the task didn't start, [Final task status] should be unknown.
5. Recieve the response of 'I have succeeded in the task: xxx' in [Conversation] means that the task has finished, [Final task status] should be a success.
6. Recieve the response of 'I have failed the task: xxx' in [Conversation] means that the task has failed, [Final task status] should fail.
5. The [Final task status] should be one of status in 'success' or 'fail' or 'unknown'
'unknown' means there is no information about the success or failure in progress analysis,
'success' means the task is finally completed,
'fail' means the task is finally failed.

The response format should be:
Task result judgment: <Your output [Task result judgment] should be like, "According to [the key sentence] from [Conversation or Chest information], [Task to be inquired] has succeeded/failed." Or "[information summary], but it does not provide any information about the result of the task, so the task status is unknown.">
Final task status: <Only one of the words in success or fail or unknown>

\end{lstlisting}
\end{tcolorbox}

\begin{tcolorbox}[title={\textbf{\small Monitor progress: example of root agent}}, colback=whitesmoke, colframe=bluegray, boxrule=2pt, arc=0mm, breakable]
{\scriptsize
\begin{lstlisting}[style=mystyle]
Example:
Input: (If you are the leader)
The current task to be inquired:```Stage 1: Gather resources
    WorkerA mine 27 stones.```
The conversation between worker and leader
-[15:03:10]leader says: 'WorkerA, please mine 27 stones'
-[15:03:20] WorkerA says: 'Got it!'
Task result judgment: Leader informs workers to mine 27 logs, workers receive it, but don't hear the message that WorkerA has succeeded, so the task status is unknown
Final task status: unknown
\end{lstlisting}}
\end{tcolorbox}

\begin{tcolorbox}[title={\textbf{\small Progress monitor: example of leaf agent}}, colback=whitesmoke, colframe=bluegray, boxrule=2pt, arc=0mm, breakable]
{\scriptsize
\begin{lstlisting}[style=mystyle]
Example:

Input:
The current task to be inquired: Craft a wooden pickaxe
Conversation:  {'linnea3v3': ["[19:35:09]workera says: 'I'll start the task Craft a wooden pickaxe now'", "[19:36:11]workera says: 'I have succeeded the task Craft a wooden pickaxe.'", "[19:36:11]workera says: 'The critique is Successfully crafted a wooden pickaxe.'", "[19:36:11]workera says: 'my inventory is {'acacia_log': 11}, and my equipment is [None, None, None, None, None, None] '"]}
Supplies in the chest: none
Output:
Task result judgment: According to the conversation, workers' inventory is {'acacia_log': 11}, and my equipment is none, so workera has failed the task Craft a wooden pickaxe.
Final task status: failed

Input:
The current task to be inquired: WorkerA mine 15 more irons. WorkerA mine 10 logs
{'linnea3v3': ["[13:49:58]workera says: 'I have failed the task mine 15 irons.'", "[13:51:55]workera says: 'I have succeeded the task mine 10 logs.'", "[13:51:55]workera says: 'my inventory is {'crafting_table': 1, 'oak_planks': 8, 'stick': 8, 'oak_log': 5, 'birch_log': 5}, and my equipment is [None, None, None, None, 'crafting_table', None] '"]}
Output:
Task result judgment: According to the inventory, workera has succeeded in the task mine 10 logs, but has failed the task mine 15 irons. Therefore, he failed.
Final task status: failed
\end{lstlisting}}
\end{tcolorbox}

\subsection{Task planner}

\subsubsection{Components in the Prompt}
The input prompt to LLM consists of several components:
\begin{enumerate}[label=(\arabic*)]
\item Recent conversations. To optimize token usage, the conversation is truncated to only include the dialogue from the last task planning session to the present moment.
\item Objective proposed by the previous task planner.
\item Long-term plan broken down by objectives, also proposed by the previous task planner. 
\item The status of the current task. The completion of the current task, proposed by the progress monitor.
\end{enumerate}
\subsubsection{Full Prompt}

\begin{tcolorbox}[title={\textbf{\small Task planner of root agent}}, colback=whitesmoke, colframe=bluegray, boxrule=2pt, arc=0mm, breakable]
{\scriptsize
\begin{lstlisting}[style=mystyle]
You are a Minecraft planner, your name is {name}, and you need to follow the rules of Minecraft and split the tasks into multiple stages to complete them according to the tasks in the [Conversation].
I will give you a [Conversation] and a [Previous long-term plan] (if have one) and a [Previous inventory of employers], and you need to output the [Current inventory of employers], [Analysis], [Long term plan], [Task at hand] and [Informer].
Your task is to perform the following actions to help me play Minecraft:
1. Handling the conversation and [Previous inventory of employers], you need to summarize the [Current inventory of employers]. Subject to the latest time of conversation
2. If there is no given [Previous long-term plan], develop one based on the conversation and [Current inventory of employers], and if there is a [Previous long-term plan], adjust it based on the conversation and [Current inventory of employers], (if there is not too much information in the [Conversation], output the original [Previous long term plan] content)
3. Then, You need to break down the tasks in conversation into step-by-step that can be performed in Minecraft, considering your social role and suppose you have nothing in your inventory.
4. Finally, please output the task at hand and who informs you to do the task.
You must follow the following criteria:
1. You now have employments {employment} in Minecraft to dispatch and divide up, and you need to try to arrange for everyone ({employment}) to have something to do in each STEP to maximize parallelism and efficiency, and not have employees waiting for and blocking each other. Don't leave your employees with nothing to do.
2. You need to consider the interdependence of tasks in Minecraft for the division of labor.
3. You need to consider {employment}'s inventory status from the conversation and develop a reasonable plan that fits their available resources (They can not get or use items from others).
4. If a [Previous long-term plan] is already in place, you need to adjust it based on their inventory and [previous progress] to reach your goals
5. You should take tasks that have a large workload and split them into smaller tasks, for example, if you need to mine 50 logs, you can split them into workerA mine 25 logs, workerB mine 25 logs.
6. You should have all your employees doing one thing and one thing only at every stage, no more and no less.
7. You have to adjust the plan so that the goal is accomplished in the shortest possible time, for example, by assigning work to an employee with a successful track record (once an employee has completed a job, assign it to someone else who hasn't) or by swapping the employee's work in hand.
8. Things that affect each other should be done in different stages (e.g. Building foundation should be before building walls, building walls should before building roof and they(building foundation, walls, roof) should be in different stages)
9. Things that don't affect each other can be done in one stage by different employees (e.g. Installing the door and building the roof do not interfere with each other and can be scheduled in the same stage for different employees to do). Please try to improve the overall efficiency as much as possible. 
10. Each step related to construction needs to contain very detailed location information.
11. You can't generate 'assistance' tasks, each employee should do different things depending on their inventory.

The response format should be:
Current inventory of employers: <Updated inventory based on a previous inventory of employers and conversation.>
Objective: <combine the conversation and the objective from the last PLAN to give the current overall objective, possibly keeping it the same>
Analysis: <the step-by-step analysis of the conversation and previous progress (if have), then decide how to plan the goal and division of labor in Minecraft based on inventory of employers>
Long-term plan: The overall plan with multiple [stage], in each stage, everyone has important and specific tasks to do
The task at hand: The [stage] needs to be done now, in this stage, everyone has a specific task. (Please answer in detailed text), Output None if there is no task need to do now.
Informer is <one of the player's names, who inform you to do this thing, you can find in conversation>
Examples:
Example 1:
[INPUT]:
Conversation: ``` The conversation between linnea3v3 and leader
-[15:33:35]linnea3v3 says: build a house begin at (-10,72,-30)
```
previous conversation:```None```
previous long-term plan: ```None```
Current inventory of employers: workerA has an empty inventory, workerB has an empty inventory, workerC has an empty inventory.

[Output]:
Current inventory of employers: workerA has an empty inventory, workerB has an empty inventory, workerC has an empty inventory.

Objective:
Build a house.

Analysis: 
To build a house, we need to break down the tasks into several stages, including gathering materials, preparing the land, laying the foundation, constructing walls, and adding the roof and finishing touches. Since we're starting from scratch with empty inventories, the first step will be to gather necessary resources like wood and stone.

Long-term plan:
Stage 1: WorkerA, WorkerB, WorkerC gather resources
    WorkerA mine 12 woods.
    WorkerB mine 25 stone.
    WorkerC mine 25 woods.
Stage 2: WorkerA builds the foundation at (-10,72,-30) WorkerB and WorkerC dig the wood 
    WorkerA builds the foundation at (-10,72,-30)
Stage 3: WorkerB builds the wood wall at (-10,72,-30) 
...
Stage 4: WorkerC builds the wood roof at (-10,72,-30) 
...

The task at hand:
Stage 1: WorkerA,WorkerB,WorkerC gather resources
    WorkerA mine 25 woods.
    WorkerB mine 15 stone.
    WorkerC mine 25 woods.

Informer is Linnea3v3


Example 2:
[INPUT]:
Conversation: ``` The conversation between linnea3v3 and leader
-[15:33:35]linnea3v3 says: mine 50 stones
```
previous conversation:```None```
previous long-term plan: ```None```
Current inventory of employers: workerA has an empty inventory, workerB has an empty inventory, workerC has an empty inventory.

[Output]:
Current inventory:
workerA has an empty inventory, workerB has an empty inventory, workerC has an empty inventory.

Objective:
Mine 50 stones.

Analysis:
To efficiently complete the task of mining 50 stones, the workload needs to be divided among workerA, workerB, and workerC. Since the task is straightforward and all workers have the same starting point (empty inventory), the task can be evenly distributed.

Long term plan:
Stage 1: Gather stones

    WorkerA mines 17 stones.
    WorkerB mines 17 stones.
    WorkerC mines 16 stones.

The task at hand:
Stage 1: Gather stones

    WorkerA mines 17 stones.
    WorkerB mines 17 stones.
    WorkerC mines 16 stones.

Informer is Linnea3v3

Example3:
[INPUT]:
Conversation: ''' The conversation between linnea3v3 and leader
-[19:15:13]linnea3v3 says: 'Mine 50 logs'
...

The conversation between workera and leader
-[19:15:37]leader says: 'WorkerA, please mine 17 logs'
-[19:17:33]workera says: 'My inventory is ['birch_log': 17, 'birch_planks': 2], and my equipment is [None, None, None, None, None, None] '


The conversation between workerb and leader
-[19:15:42]leader says: 'workerB, please mine 17 logs'
-[19:16:09]workerb says: 'I'll start the task mine 17 logs now'

The conversation between workerc and the leader
-[19:15:48]leader says: 'workerC, please mine 16 logs'
-[19:16:19]workerc says: 'I'll start the task mine 16 logs now'

'''
Previous objective: '''Mine 50 logs.'''
Previous long term plan: '''Stage 1: Gather logs

    WorkerA mines 17 logs.
    WorkerB mines 17 logs.
    WorkerC mines 16 logs.'''
The previous progress we have done is the step Stage 1: Gather stones

    WorkerA mines 17 logs.
    WorkerB mines 17 logs.
    WorkerC mines 16 logs. is failed

[Output]:
Current inventory of employers: ...
Objective:
Complete the task of mining 50 stones.

Analysis:
A has successfully completed the task of mining 17 logs, however, b and c have not, to maximize time efficiency try to get A to mine the remaining 33 logs and B and C to continue mining

Long term plan:
Stage 1 (adjust plan):
    WorkerA mined the remaining 33 logs.
    WorkerB mines 17 logs

The task at hand:
Stage 1 (adjust plan):
    WorkerA mined the remaining 33 logs.
    WorkerB mines 17 logs

Informer is Linnea3v3

\end{lstlisting}}
\end{tcolorbox}

\begin{tcolorbox}[title={\textbf{\small Task planner of Leaf agent}}, colback=whitesmoke, colframe=bluegray, boxrule=2pt, arc=0mm, breakable]
\scriptsize
{\begin{lstlisting}[style=mystyle]
You are a Minecraft planner, and you need to follow the rules of Minecraft and split the tasks into multiple stages to complete them according to the tasks in the [Conversation].
Your task is to perform the following actions to help me play Minecraft:
1. Your name is {name}. Handling the conversation and your inventory status. If there is [previous progress], you need to consider it when you make a plan.
2. If there is no given [Previous long-term plan], develop one based on the conversation, and if there is a [Previous long-term plan], adjust it based on the conversation, (if there is not too much information in the [Conversation], output the original [Previous long term plan] content)
3. You need to break down the tasks in conversation step-by-step that can be performed in Minecraft, considering your social role and suppose you have nothing in your inventory.
4. Then, please output the task at hand and who informs you to do the task.
You must follow the following criteria:
1. Since you are a WORKER, you have no Employment{employment}. you receive the plan in the chat logs and you should do it yourself.
2. You need to consider your inventory status and make a plan that is easy to succeed in Minecraft.
3. Please do not build shelters without authorization, but if you are asked to build something, you should follow orders.
4. When mining logs, please do not craft a wooden axe, you should simply mine wood by hand. But, if someone asks you to craft an axe, you should do so.
5. When mining stone, you should first mine logs, then craft and equip a pickaxe (e.g. a wooden pickaxe).
6. When collecting resources such as Log and Stone, please collect more than one at a time (e.g. 10 or more) just in case!
7. If the previous plan failed due to a timeout error, please try again with the same plan, do not change the plan.
8. If you already have more than the required amount of supplies in your inventory, you don't need to generate mine tasks!
9. Each step related to construction needs to contain very detailed location information. Each building step should contain location information.

The response format should be:
Current inventory: <current inventory, inferred from the conversation and inventory>
Objective: <combine the conversation and the objective from the last PLAN to give the current overall objective, possibly keeping it the same>
Analysis: <the step-by-step analysis of the conversation and previous progress (if have), then decide how to plan the goal in Minecraft>
Long term plan: The overall plan with multiple [stage], in each stage, everyone has important and specific tasks to do
The task at hand: The [stage] needs to be done now, in this stage, everyone has a specific task. (Please answer in detailed text), Output None if there is no task need to do now.
Informer is <one of the player's names, who inform you to do this thing, you can find in conversation>
Examples:
Example 1:
Input:
Conversation: ``` The conversation between leader and workerA
-[15:33:35]leader says: WorkerA mine 10 woods.```
previous objective:```None```
previous long term plan: ```None```

Output:
Current inventory: 
the inventory of workerA is None

Objective:
mine 10 woods.

Analysis:  
To mine 10 woods in Minecraft, you need to start by finding trees. Since you're starting with nothing in your inventory, you'll have to mine the wood by hand.

Long term plan:
Stage 1: WorkerA gathers resources
    WorkerA mine 10 woods.
The task at hand:
    WorkerA mine 10 woods.
Informer is leader


Example2:
Input:
Conversation: ``` The conversation between leader and workerA
-[15:33:35]leader says: WorkerA mine 25 stones.```
previous objective:```None```
previous long term plan: ```None```

Output:
Current inventory:
the inventory of workerA is None

Objective:
mine 25 stones.

Analysis:
To mine 25 stones in Minecraft, you must first gather wood to craft wooden pickaxes, as mining stones directly by hand won't yield any resources. Start by finding and mining at least 3 logs from trees. Then, use a crafting table to convert these logs into wooden planks and sticks. With these materials, craft a wooden pickaxe to start mining stone.

Long term plan:
Stage 1: Gather resources

    WorkerA mine 3 logs.
    WorkerA craft wooden planks and sticks from logs.
    WorkerA crafts a wooden pickaxe.

Stage 2: Mine stone

    WorkerA uses the wooden pickaxe to mine 25 stones.

The task at hand:
Stage 1: Gather resources

    WorkerA mine 3 logs.
    WorkerA craft wooden planks and sticks from logs.
    WorkerA crafts a wooden pickaxe.

Informer is leader

\end{lstlisting}}
\end{tcolorbox}

\subsection{Action planner}

\subsubsection{Components in the Prompt}
The input prompt to llm us the current task proposed by the task planner.

\subsubsection{Full Prompt}
\begin{tcolorbox}[title={\textbf{\small Template of action planner}}, colback=whitesmoke, colframe=bluegray, boxrule=2pt, arc=0mm, breakable]
\scriptsize
\begin{lstlisting}[style=mystyle]
You are a helpful assistant. Your task is to directly translate [Current task] input into [TODO list] as RESPONSE. Your [TODO list] translations must be consistent with the [Current task], especially the 'who does each task' issue. 
1. - You are player {name}. Your profile is ```{profile}```.
2. - [TODO list] has two types of items, those that you do yourself (e.g. Craft [quantity] [item] (at position)) and those that you arrange for someone else to do (e.g. inform [player] to kill [quantity] [mob] (at position))
3. - If the [current task] is someone else's, you should not translate the task you did yourself in [TODO list], but you should INFORM that person!

You must follow the following criteria:
1. The [todo list] items should follow a concise format, such as "Mine [quantity] [block] (at position)", "inform [player] to mine [quantity] [block] (at position)", "inform [player] to kill [quantity] [mob] (at position)", "Craft [quantity] [item] (at position)", "Smelt [quantity] [item] (at position)", "Kill [quantity] [mob] (at position)", "Cook [quantity] [food] (at position)", "Equip [item] (at position)", "Build [item] at [position]" etc. It should be a single phrase. Do not mention anything else. mention position when necessary.
2. The [to-do list] items should have an exact position if there is position information in the conversation. 
3. When the task contains someone else's name ({employment}), you need to generate the inform todo, like "inform [player] to mine [quantity] [block] (at position)", "inform [player] to kill [quantity] [mob] (at position)", "inform [player1] to give [quantity] [item] to [player2]".

Response format should be:
<a list of todo, like ["todo1", "todo2", "todo3", ...]>(This JSON format will be parsed by Python `json.loads`)
Ensure the response can be parsed by Python `json.loads`, e.g.: no trailing commas, no single quotes, etc.


EXAMPLE:
{example}
\end{lstlisting}
\end{tcolorbox}

\begin{tcolorbox}[title={\textbf{\small Action planner: example of root agent}}, colback=whitesmoke, colframe=bluegray, boxrule=2pt, arc=0mm, breakable]
\scriptsize
\begin{lstlisting}[style=mystyle]
Example 1: If your name is leader, you should better generate task that includes 'inform':
INPUT:
Current task:
Step: gather resources
    WorkerA mine 25 woods.
    WorkerB mine 15 stone.
RESPONSE:
["inform WorkerA to mine 25 woods", "inform workerB mine 15 stone"]

Example 2: If your name is leader:
INPUT:
Current task: 
Step: WorkerA needs to build the foundation.
RESPONSE:
["inform workerA to build the foundation"]

Example 3:
INPUT:
Current task:
Stage 2: WorkerA builds the walls
    WorkerA uses 48 planks to build the walls at (-10,72,-30).
RESPONSE:
["inform workerA to build walls at (-10,72,-30) (use 48 planks)"]

Example 4:
INPUT:
Current task:
Stage 1: WorkerA crafts 4 planks from the log of WorkerB.

RESPONSE:
["inform workerB to give a log to workerA", "inform workerA to craft 4 planks"]

\end{lstlisting}
\end{tcolorbox}

\begin{tcolorbox}[title={\textbf{\small Action planner: example of leaf agent}}, colback=whitesmoke, colframe=bluegray, boxrule=2pt, arc=0mm, breakable]
\scriptsize
\begin{lstlisting}[style=mystyle]
Example 1: If your name is worker, you should not generate a task includes 'inform', you should only do the task yourself:
INPUT:
Current task:
Step: WorkerA mine 25 woods
RESPONSE:
["mine 25 woods"]
\end{lstlisting}
\end{tcolorbox}

\end{document}